\documentclass{article}

\usepackage{PRIMEarxiv}
\usepackage{amsmath}
\numberwithin{equation}{subsection}
\usepackage{amsthm}
\usepackage{amssymb}
\usepackage{mathtools}
\usepackage{enumitem}
\usepackage{adjustbox}
\usepackage{booktabs}
\usepackage{multirow}
\usepackage[utf8]{inputenc}
\usepackage[T1]{fontenc}
\usepackage{hyperref}
\usepackage{url}
\usepackage{booktabs}
\usepackage{amsfonts}
\usepackage{nicefrac}
\usepackage{microtype}
\usepackage{lipsum}
\usepackage{fancyhdr}
\usepackage{graphicx}
\usepackage{natbib}
\graphicspath{{media/}}

\newtheorem{theorem}{Theorem}

\newtheorem{proposition}[theorem]{Proposition}
\newtheorem{corollary}[theorem]{Corollary}

\newtheorem{definition}{Definition}

\title{A Ridge Too Far: Correcting Over-Shrinkage via Negative Regularization}

\author{
  Dongseok Kim, Gisung Oh \\
  Department of Computer Engineering \\
  Gachon University \\
  Seongnam, Gyeonggi, Republic of Korea \\
  \texttt{\{jkds5920, eustia\}@gachon.ac.kr}
}

\begin{document}
\maketitle

\begin{abstract}
Conventional regularization is designed to control variance, but in small-data regression it can also aggravate underfitting when predictive signal is concentrated in weak directions of a restricted representation. We study a negative-capable ridge family that permits a feasible negative region whenever the estimator remains well posed, and show that negative regularization acts there as controlled anti-shrinkage by increasing effective complexity most strongly along weak eigendirections. Building on this mechanism, we formalize weak-spectrum underfitting, derive a sign-switch result under conservative baseline shrinkage, and study criterion-based automatic selection over the full negative-capable family. Synthetic and semi-synthetic experiments support the theory by verifying feasibility, spectral complexity increase, sign-switch behavior, and effective recovery of negative adjustments in the predicted regimes.
\end{abstract}


\section{Introduction}
\label{sec:introduction}

Regularization is usually introduced as a device for controlling variance and preventing overfitting. In many learning problems, this intuition is appropriate: when the predictor is too flexible relative to the available sample size, shrinking the estimator toward a simpler solution can improve generalization. However, this standard perspective becomes incomplete in small-data settings where prediction is performed through a restricted representation. In such regimes, the dominant difficulty may be not excessive flexibility but excessive conservativeness. When the available representation already imposes a bottleneck, additional shrinkage can suppress precisely those directions that remain necessary for accurate prediction within the restricted space.

This paper studies that regime in the context of small-data regression under a fixed representation bottleneck. Our starting point is that underfitting should not be viewed only as a generic lack of model capacity. It can also arise as a structural spectral mismatch between the geometry of the predictive signal and the geometry induced by regularization. In particular, when the restricted oracle places a nontrivial portion of its mass in weak eigendirections of the representation covariance, standard nonnegative regularization can amplify bias by attenuating exactly those coordinates that are already fragile. From this viewpoint, the relevant question is not simply how much regularization to add, but whether the current shrinkage pattern is itself part of the underfitting problem.

Motivated by this observation, we study a negative-capable regularization family that allows the regularization parameter to enter a feasible negative region whenever the optimization problem remains well posed. The point of this formulation is not to advocate instability or unrestricted complexity growth. Rather, the negative region is interpreted as a form of controlled anti-shrinkage: it relaxes an already conservative estimator and can partially reverse excessive shrinkage bias while remaining inside an explicit spectral admissibility range. This makes negative regularization meaningful not as an exceptional trick, but as a principled correction mechanism for a specific small-data underfitting structure.

The central theoretical issue is then a sign question: under what structural conditions should the optimal adjustment move into the negative region? To answer this, we show that the effect of negative regularization is spectrally ordered, increasing effective complexity most strongly along weak eigendirections. We then formalize a weak-spectrum underfitting regime and prove that, relative to a conservative baseline, the oracle preference can switch sign: when bias reduction dominates the accompanying variance inflation, a negative adjustment becomes optimal. Since such a result is meaningful only if the negative region can also be used in practice, we further study data-driven selection over the full negative-capable family through a risk-estimation criterion and analyze when the selected parameter can recover the benefit of the negative regime.

Our contributions are as follows:
\begin{itemize}
    \item \textbf{A structural spectral view of small-data underfitting.} Rather than treating underfitting as a purely global shortage of flexibility, we identify a regime in which the restricted oracle is aligned with weak eigendirections of the representation covariance, so that ordinary shrinkage suppresses the coordinates that matter most within the available representation space.

    \item \textbf{A negative-capable regularization framework with a feasible negative interval.} We characterize the range in which negative regularization remains well posed, and we show that moving into this region acts as controlled anti-shrinkage by increasing effective complexity in a directionally structured way rather than through indiscriminate destabilization.

    \item \textbf{A sign-switch theory for conservative small-data learning.} Relative to a positive baseline shrinkage level, we derive conditions under which the oracle adjustment becomes negative. The result makes explicit that negative regularization is not universally preferable, but becomes justified when shrinkage-induced bias in weak-spectrum directions outweighs the resulting variance increase.

    \item \textbf{A criterion-based automatic selection method over the full negative-capable family.} Using a data-driven selector, we analyze how one can search over both negative and nonnegative adjustments within the feasible region, and we provide guarantees linking the selected parameter to the best candidate in the family.

    \item \textbf{An empirical verification of the proposed mechanism.} Through synthetic and semi-synthetic regression experiments, we examine feasibility of the negative region, effective complexity increase, weak-spectrum bias concentration, oracle sign-switch behavior, and automatic selection performance, showing that the advantage of negative adjustment appears in the structurally predicted regimes rather than uniformly.
\end{itemize}

\section{Related Work}
\label{sec:related_work}

\subsection{Ridge Regression and Shrinkage}
\label{subsec:rw_ridge_shrinkage}

Ridge regression originated as a biased estimation strategy for ill-conditioned linear models, and the classical literature established its basic multicollinearity motivation, mean-squared-error justification, data-driven tuning, and generalized minimax variants~\citep{hoerl1970ridge, theobald1974generalizations, golub1979generalized, strawderman1978minimax, maruyama2005new}. The broader shrinkage literature then extended this line of work beyond isotropic $\ell_2$ penalization to coefficient garroting, $\ell_1$ and bridge penalties, path-following algorithms, mixed $\ell_1/\ell_2$ regularization, adaptive weighting, grouped selection, fused penalties for ordered features, and global-local Bayesian shrinkage~\citep{breiman1995better, tibshirani1996regression, efron2004least, fu1998penalized, zou2005regularization, zou2006adaptive, yuan2006model, tibshirani2005sparsity, carvalho2010horseshoe}. Related developments also carried ridge-style regularization into kernel methods with explicit statistical guarantees~\citep{alaoui2015fast}.

\subsection{Restricted Representations and Underfitting}
\label{subsec:rw_restricted_underfitting}

A substantial literature studies regression through reduced or restricted representations. Classical sufficient dimension reduction work sought low-dimensional projections that preserve response-relevant structure, including sliced inverse regression, conditional-mean formulations, and broader syntheses of regression dimension reduction~\citep{li1991sliced, cook2002dimension, cook2007fisher, cook2008principal, cook2005sufficient}. Subsequent developments introduced directional, constructive, kernel, and nonlinear extensions, together with estimators and asymptotic analyses tailored to sufficient reductions in high-dimensional settings~\citep{li2007directional, xia2007constructive, fukumizu2009kernel, li2011principal, cook2012estimating, lee2013general}. In parallel, related work in statistics and machine learning considered supervised low-dimensional summaries and explicit finite feature restrictions, including supervised principal components and randomized kernel feature maps that replace richer function classes with tractable compressed representations~\citep{bair2006prediction, rahimi2007random, rahimi2008weighted, le2013fastfood}.

\subsection{Negative Regularization}
\label{subsec:rw_negative_regularization}

Recent top-venue work has shown that the optimal amount of ridge regularization need not be strictly positive, and can become zero or negative depending on feature anisotropy, signal alignment, implicit shrinkage, and data geometry~\citep{hsu2012random, dobriban2018high, kobak2020optimal, wu2020optimal, tsigler2023benign}. Closely related analyses of ridgeless and near-ridgeless regimes studied when interpolation or near-interpolation remains statistically favorable, both in finite-feature linear models and in kernel or structured-feature settings~\citep{liang2020just, richards2021asymptotics, mcrae2022harmless, hastie2022surprises, wang2024near}. Another line of work examined how such nonclassical regularization regimes can be selected or characterized under broader testing conditions, including cross-validation over ranges that include negative values, out-of-distribution prediction, correlated-sample risk estimation, robustness-oriented analyses, and convolutional spectral models~\citep{patil2021uniform, patil2024optimal, atanasov2024risk, donhauser2021interpolation, sahraee2021asymptotics}.

\subsection{Bias--Variance Trade-offs and Sign-Switch Behavior}
\label{subsec:rw_bias_variance_sign_switch}

A broad literature analyzes generalization through bias--variance decompositions, complexity measures, and related diversity decompositions in kernels, ensembles, and augmentation settings~\citep{valentini2004bias, wood2023unified, lee2020neural, chen2020group}. More recent work questioned the classical monotone picture by showing that test risk can exhibit double-descent or other non-monotone regime changes as model size, sample size, or effective complexity varies~\citep{belkin2019reconciling, nakkiran2021deep, yang2020rethinking, d2020double, d2020triple}. Related analyses further decomposed error beyond the basic bias--variance split, showing how initialization, label noise, uncertainty criteria, or principal-component truncation can alter whether complexity increases help or hurt~\citep{lin2021causes, hodgkinson2023monotonicity, gedon2024no}. A complementary line of work studied early stopping and boosting as pathwise or iterative regularization mechanisms, emphasizing that the effective bias--variance balance can change with stopping time and along the optimization path~\citep{raskutti2014early, wei2017early, rosset2004boosting}.

\subsection{Automatic Regularization Selection}
\label{subsec:rw_automatic_selection}

A long-standing line of work studies automatic regularization and model-complexity selection through predictive risk estimation, information criteria, and cross-validation, including foundational developments of cross-validation, Akaike's information criterion, Bayesian information criteria, and asymptotic optimality analyses for $C_p$, $C_L$, cross-validation, and generalized cross-validation~\citep{stone1974cross, akaike2003new, schwarz1978estimating, li1987asymptotic}. Related work also developed resampling-based and error-controlled calibration procedures, most notably stability selection and its refinements for structured high-dimensional selection~\citep{meinshausen2010stability, shah2013variable}. In sparse regression, automatic tuning has been analyzed through degrees-of-freedom and Stein-based risk estimation, consistency results for LASSO tuning, and cross-validated high-dimensional lasso theory~\citep{zou2007degrees, tibshirani2012degrees, mousavi2018consistent, chetverikov2021cross, bellec2021second}. More recent machine learning work studies efficient hyperparameter optimization and approximate validation schemes for nonsmooth or structured regularized estimators, including bilevel differentiation methods, direct $\ell_p$ hyperparameter learning, iterative approximate cross-validation, and adaptive tuning for graphical lasso procedures~\citep{bertrand2022implicit, okuno2021lp, luo2023iterative, laszkiewicz2021thresholded}.

\section{Theory}
\label{sec:theory}

\subsection{Problem Setup}
\label{subsec:theory_setup}

We consider a small-data regression problem in which the predictor is learned not in an unrestricted function class, but in a restricted representation space. Let $(X,Y)$ be a random pair with $X \in \mathcal{X}$ and $Y \in \mathbb{R}$ satisfying
\[
Y = f_\star(X) + \varepsilon,
\qquad
\mathbb{E}[\varepsilon \mid X] = 0,
\qquad
\mathbb{E}[\varepsilon^2 \mid X] = \sigma^2,
\]
where $f_\star : \mathcal{X} \to \mathbb{R}$ is the unknown regression function. We observe an i.i.d.\ sample
\[
\mathcal{D}_n = \{(x_i,y_i)\}_{i=1}^n .
\]

We assume that learning is performed through a fixed $m$-dimensional representation map
\[
\phi : \mathcal{X} \to \mathbb{R}^m,
\qquad
\phi(x) = (\phi_1(x),\dots,\phi_m(x))^\top,
\]
and we restrict attention to linear predictors of the form
\[
f_\beta(x) = \phi(x)^\top \beta,
\qquad
\beta \in \mathbb{R}^m .
\]
This restricted representation is the source of the structural underfitting considered in this paper: even when $f_\star$ is not itself linear in $\phi(x)$, the learner is forced to approximate it within the span of the chosen coordinates.

Let
\[
\Sigma := \mathbb{E}[\phi(X)\phi(X)^\top]
\]
denote the population covariance of the representation, and let
\[
\hat{\Sigma}_n := \frac{1}{n}\sum_{i=1}^n \phi(x_i)\phi(x_i)^\top
\]
be its empirical counterpart. We also define the population cross-moment
\[
g := \mathbb{E}[\phi(X)Y] = \mathbb{E}[\phi(X)f_\star(X)] .
\]
Throughout the theory section, we assume that $\Sigma$ is symmetric positive semidefinite and that the second moments of $\phi(X)$ and $Y$ are finite.

The population-optimal coefficient within the restricted representation space is defined by
\[
\beta^\dagger \in \arg\min_{\beta \in \mathbb{R}^m}
\mathbb{E}\bigl[(Y-\phi(X)^\top \beta)^2\bigr].
\]
Whenever $\Sigma$ is invertible, this target is unique and satisfies
\[
\beta^\dagger = \Sigma^{-1} g.
\]
More generally, if $\Sigma$ is singular, $\beta^\dagger$ may be taken as the minimum-norm solution of the normal equation
\[
\Sigma \beta = g.
\]
Thus, $\beta^\dagger$ is not the coefficient of the true regression function itself, but the best $L_2(P_X)$ approximation to $f_\star$ within the restricted representation family.

Given the sample $\mathcal{D}_n$, we study the negative-capable ridge family
\[
\hat{\beta}_\lambda
\in
\arg\min_{\beta \in \mathbb{R}^m}
\left\{
\frac{1}{n}\sum_{i=1}^n (y_i-\phi(x_i)^\top \beta)^2
+
\lambda \|\beta\|_2^2
\right\},
\]
where $\lambda \in \mathbb{R}$ is allowed to be negative whenever the optimization problem remains well-posed. Writing
\[
\Phi :=
\begin{bmatrix}
\phi(x_1)^\top \\
\vdots \\
\phi(x_n)^\top
\end{bmatrix}
\in \mathbb{R}^{n \times m},
\qquad
Y :=
\begin{bmatrix}
y_1 \\
\vdots \\
y_n
\end{bmatrix}
\in \mathbb{R}^n,
\]
the estimator takes the formal closed form
\[
\hat{\beta}_\lambda
=
(\hat{\Sigma}_n + \lambda I_m)^{-1}\hat{g}_n,
\qquad
\hat{g}_n := \frac{1}{n}\Phi^\top Y,
\]
whenever $\hat{\Sigma}_n + \lambda I_m$ is invertible. The precise admissible negative range will be characterized in the next subsection.

Our target of analysis is the population prediction risk
\[
\mathcal{R}(\beta)
:=
\mathbb{E}\bigl[(Y-\phi(X)^\top \beta)^2\bigr],
\]
and, in particular, the excess risk relative to the restricted oracle $\beta^\dagger$,
\[
\mathcal{E}(\beta)
:=
\mathcal{R}(\beta) - \mathcal{R}(\beta^\dagger).
\]
This excess-risk formulation is natural in the present setting because the representation bottleneck may prevent the learner from attaining the Bayes risk even with infinite data. The quantity $\mathcal{E}(\beta)$ isolates the part of the error attributable to estimation and regularization inside the restricted model family.

The following proposition gives the basic quadratic form of the excess risk and identifies $\beta^\dagger$ as the orthogonal projection target in the geometry induced by $\Sigma$.

\begin{proposition}
\label{prop:excess_risk_quadratic}
For every $\beta \in \mathbb{R}^m$,
\[
\mathcal{E}(\beta)
=
(\beta-\beta^\dagger)^\top \Sigma (\beta-\beta^\dagger).
\]
Equivalently,
\[
\mathcal{R}(\beta)
=
\mathcal{R}(\beta^\dagger)
+
\|\Sigma^{1/2}(\beta-\beta^\dagger)\|_2^2 .
\]
In particular, $\beta^\dagger$ is a minimizer of $\mathcal{R}(\beta)$ over $\mathbb{R}^m$, and it is unique whenever $\Sigma$ is positive definite.
\end{proposition}

The proposition identifies $\beta^\dagger$ as the natural oracle target inside the restricted representation family and shows that excess risk is measured exactly by the quadratic geometry induced by $\Sigma$. The proof is deferred to Appendix~\ref{app:proofs_feasible_complexity}.

To prepare for the sign-switch analysis, it is useful to introduce the population regularized target
\[
\beta_\lambda^{\mathrm{pop}}
\in
\arg\min_{\beta \in \mathbb{R}^m}
\left\{
\mathcal{R}(\beta) + \lambda \|\beta\|_2^2
\right\}.
\]
When $\Sigma + \lambda I_m$ is invertible, this target satisfies
\[
\beta_\lambda^{\mathrm{pop}}
=
(\Sigma+\lambda I_m)^{-1} g
=
(\Sigma+\lambda I_m)^{-1}\Sigma \beta^\dagger .
\]

\subsection{Feasible Negative Interval}
\label{subsec:theory_feasible_interval}

Negative regularization is meaningful only when the associated quadratic objective remains bounded below and admits a unique minimizer. We therefore begin by characterizing the admissible negative region spectrally. Throughout this subsection, write the empirical criterion as
\[
Q_n(\beta;\lambda)
:=
\frac{1}{n}\|Y-\Phi\beta\|_2^2+\lambda\|\beta\|_2^2
=
\frac{1}{n}\|Y\|_2^2
-2\hat g_n^\top \beta
+
\beta^\top(\hat\Sigma_n+\lambda I_m)\beta,
\]
where
\[
\hat\Sigma_n=\frac{1}{n}\Phi^\top\Phi,
\qquad
\hat g_n=\frac{1}{n}\Phi^\top Y .
\]
Let
\[
\hat\mu_1 \ge \hat\mu_2 \ge \cdots \ge \hat\mu_m \ge 0
\]
denote the eigenvalues of $\hat\Sigma_n$, and let
\[
\mu_1 \ge \mu_2 \ge \cdots \ge \mu_m \ge 0
\]
denote the eigenvalues of the population covariance matrix $\Sigma$. For the main theory, we focus on the nondegenerate regime in which the working representation is identifiable at the population level, namely $\Sigma \succ 0$.

\begin{proposition}[Sample well-posedness]
\label{prop:sample_well_posedness}
For a given $\lambda \in \mathbb{R}$, the following statements are equivalent:
\begin{enumerate}
    \item[(i)] The function $\beta \mapsto Q_n(\beta;\lambda)$ is bounded below on $\mathbb{R}^m$ and admits a unique minimizer.
    \item[(ii)] The matrix $\hat\Sigma_n+\lambda I_m$ is positive definite.
    \item[(iii)] $\lambda > -\hat\mu_m$.
\end{enumerate}
Whenever these conditions hold,
\[
\hat\beta_\lambda
=
(\hat\Sigma_n+\lambda I_m)^{-1}\hat g_n
\]
is the unique minimizer, and
\[
\bigl\|(\hat\Sigma_n+\lambda I_m)^{-1}\bigr\|_{\mathrm{op}}
=
\frac{1}{\hat\mu_m+\lambda},
\qquad
\|\hat\beta_\lambda\|_2
\le
\frac{\|\hat g_n\|_2}{\hat\mu_m+\lambda}.
\]
\end{proposition}

The proposition shows that negative regularization is admissible only to the left of $0$ but to the right of the empirical spectral boundary $-\hat\mu_m$. This gives the precise sense in which the negative region is feasible rather than arbitrary.

\begin{definition}[Empirical feasible negative interval]
\label{def:empirical_feasible_negative_interval}
Suppose $\hat\mu_m>0$. The empirical feasible negative interval is
\[
\hat\Lambda_n^{-}
:=
(-\hat\mu_m,\,0).
\]
More generally, the full empirical well-posed region is
\[
\hat\Lambda_n
:=
(-\hat\mu_m,\,\infty).
\]
\end{definition}

If $\hat\mu_m=0$, then $\hat\Lambda_n^{-}$ is empty. In other words, a nontrivial negative region exists only when the empirical Gram matrix is strictly positive definite on the working representation space.

We next record the population analogue, which will later serve as the deterministic reference for the sign-switch analysis.

\begin{proposition}[Population well-posedness]
\label{prop:population_well_posedness}
Assume $\Sigma \succ 0$ and let $\mu_m>0$ be its smallest eigenvalue. Then, for any $\lambda > -\mu_m$, the population regularized objective
\[
Q^{\mathrm{pop}}(\beta;\lambda)
:=
\mathcal{R}(\beta)+\lambda\|\beta\|_2^2
\]
is strongly convex and admits the unique minimizer
\[
\beta_\lambda^{\mathrm{pop}}
=
(\Sigma+\lambda I_m)^{-1}g
=
(\Sigma+\lambda I_m)^{-1}\Sigma \beta^\dagger .
\]
Equivalently, the population feasible negative interval is
\[
\Lambda_{\mathrm{pop}}^{-}
:=
(-\mu_m,\,0).
\]
Moreover,
\[
\bigl\|(\Sigma+\lambda I_m)^{-1}\bigr\|_{\mathrm{op}}
=
\frac{1}{\mu_m+\lambda},
\qquad
\|\beta_\lambda^{\mathrm{pop}}\|_2
\le
\frac{\|g\|_2}{\mu_m+\lambda}.
\]
\end{proposition}

The sample and population statements together make clear that the left boundary is spectral, not heuristic: the negative region ends exactly where the quadratic objective loses positive definiteness.

\begin{corollary}[Uniform interior stability]
\label{cor:uniform_interior_stability}
Fix $\tau>0$ such that $\tau<\hat\mu_m$. Define the truncated empirical feasible interval
\[
\hat\Lambda_n^{-}(\tau)
:=
[-\hat\mu_m+\tau,\,0).
\]
Then, for every $\lambda \in \hat\Lambda_n^{-}(\tau)$,
\[
\bigl\|(\hat\Sigma_n+\lambda I_m)^{-1}\bigr\|_{\mathrm{op}}
\le \frac{1}{\tau},
\qquad
\|\hat\beta_\lambda\|_2
\le \frac{\|\hat g_n\|_2}{\tau}.
\]
Moreover, for any $\lambda_1,\lambda_2 \in \hat\Lambda_n^{-}(\tau)$,
\[
\|\hat\beta_{\lambda_1}-\hat\beta_{\lambda_2}\|_2
\le
\frac{|\lambda_1-\lambda_2|}{\tau^2}\,\|\hat g_n\|_2.
\]
\end{corollary}

This corollary explains why later optimization over negative values is carried out on an interior subset separated away from the spectral boundary. Detailed proofs for Proposition~\ref{prop:sample_well_posedness}, Proposition~\ref{prop:population_well_posedness}, and Corollary~\ref{cor:uniform_interior_stability} are given in Appendix~\ref{app:proofs_feasible_complexity}.

\subsection{Effective Complexity Increase}
\label{subsec:theory_effective_complexity}

We now show that, within the feasible negative interval, decreasing $\lambda$ acts as an anti-shrinkage operation that increases the effective complexity of the estimator. The key point is spectral: negative regularization does not alter all directions uniformly, but expands coefficient recovery most strongly in the weak eigendirections that are ordinarily suppressed by positive shrinkage.

Let the population covariance matrix admit the eigendecomposition
\[
\Sigma = U \operatorname{diag}(\mu_1,\dots,\mu_m) U^\top,
\qquad
\mu_1 \ge \cdots \ge \mu_m > 0,
\]
where $U=(u_1,\dots,u_m)$ is orthogonal. Since $\beta^\dagger \in \mathbb{R}^m$, we may write
\[
\beta^\dagger = \sum_{j=1}^m \alpha_j u_j,
\qquad
\alpha_j := u_j^\top \beta^\dagger .
\]
By Proposition~\ref{prop:population_well_posedness}, for every $\lambda > -\mu_m$,
\[
\beta_\lambda^{\mathrm{pop}}
=
(\Sigma+\lambda I_m)^{-1}\Sigma \beta^\dagger .
\]

\begin{proposition}[Spectral form of the population regularized target]
\label{prop:spectral_form_population}
For every $\lambda > -\mu_m$,
\[
\beta_\lambda^{\mathrm{pop}}
=
\sum_{j=1}^m
\frac{\mu_j}{\mu_j+\lambda}\,\alpha_j u_j.
\]
Equivalently,
\[
\beta_\lambda^{\mathrm{pop}} - \beta^\dagger
=
-\sum_{j=1}^m
\frac{\lambda}{\mu_j+\lambda}\,\alpha_j u_j.
\]
In particular, the coordinate of $\beta_\lambda^{\mathrm{pop}}$ along $u_j$ is obtained by multiplying the oracle coordinate $\alpha_j$ by the shrinkage factor
\[
s_j(\lambda) := \frac{\mu_j}{\mu_j+\lambda}.
\]
\end{proposition}

The proposition shows that the regularized target is obtained by a directionwise rescaling of the restricted oracle. For $\lambda>0$, one has $0<s_j(\lambda)<1$, whereas for $\lambda<0$, one has $s_j(\lambda)>1$. Thus negative regularization acts as anti-shrinkage rather than ordinary shrinkage.

\begin{proposition}[Anti-shrinkage monotonicity]
\label{prop:anti_shrinkage_monotonicity}
Fix $j \in \{1,\dots,m\}$ and define
\[
s_j(\lambda)=\frac{\mu_j}{\mu_j+\lambda},
\qquad
\lambda > -\mu_j.
\]
Then
\[
\frac{d}{d\lambda}s_j(\lambda)
=
-\frac{\mu_j}{(\mu_j+\lambda)^2}
<
0.
\]
Therefore:
\begin{enumerate}
    \item[(i)] $s_j(0)=1$;
    \item[(ii)] if $\lambda>0$, then $0<s_j(\lambda)<1$;
    \item[(iii)] if $-\mu_j<\lambda<0$, then $s_j(\lambda)>1$;
    \item[(iv)] if $i<j$, so that $\mu_i \ge \mu_j$, then for every fixed $\lambda<0$,
    \[
    s_i(\lambda)\le s_j(\lambda),
    \]
    with strict inequality whenever $\mu_i>\mu_j$.
\end{enumerate}
\end{proposition}

Part~(iv) is the key geometric statement: the expansion is strongest on smaller eigenvalues. Negative regularization therefore does not simply increase complexity globally; it tilts recovery toward directions that ordinary shrinkage suppresses most severely.

To quantify this complexity increase at the level of predictions, define the population effective degrees of freedom by
\[
\operatorname{df}(\lambda)
:=
\operatorname{tr}\!\bigl(\Sigma(\Sigma+\lambda I_m)^{-1}\bigr),
\qquad
\lambda>-\mu_m.
\]

\begin{proposition}[Monotonic increase of effective complexity]
\label{prop:effective_df_monotonicity}
For every $\lambda>-\mu_m$,
\[
\operatorname{df}(\lambda)
=
\sum_{j=1}^m \frac{\mu_j}{\mu_j+\lambda}.
\]
Moreover,
\[
\frac{d}{d\lambda}\operatorname{df}(\lambda)
=
-\sum_{j=1}^m \frac{\mu_j}{(\mu_j+\lambda)^2}
<
0.
\]
Hence:
\begin{enumerate}
    \item[(i)] if $\lambda>0$, then $\operatorname{df}(\lambda)<m$;
    \item[(ii)] if $\lambda=0$, then $\operatorname{df}(0)=m$;
    \item[(iii)] if $-\mu_m<\lambda<0$, then $\operatorname{df}(\lambda)>m$;
    \item[(iv)] $\operatorname{df}(\lambda)\to\infty$ as $\lambda\downarrow -\mu_m$.
\end{enumerate}
\end{proposition}

The result formalizes the central mechanism of this subsection: moving leftward into the negative region increases effective complexity monotonically, and this increase becomes singular near the spectral boundary.

The same phenomenon appears at the empirical level. Define
\[
\widehat{\operatorname{df}}_n(\lambda)
:=
\operatorname{tr}\!\bigl(\hat\Sigma_n(\hat\Sigma_n+\lambda I_m)^{-1}\bigr),
\qquad
\lambda>-\hat\mu_m.
\]

\begin{proposition}[Empirical effective complexity]
\label{prop:empirical_effective_complexity}
For every $\lambda>-\hat\mu_m$,
\[
\widehat{\operatorname{df}}_n(\lambda)
=
\sum_{j=1}^m \frac{\hat\mu_j}{\hat\mu_j+\lambda},
\qquad
\frac{d}{d\lambda}\widehat{\operatorname{df}}_n(\lambda)
=
-\sum_{j=1}^m \frac{\hat\mu_j}{(\hat\mu_j+\lambda)^2}
<
0.
\]
In particular,
\[
\lambda_1<\lambda_2
\quad\Longrightarrow\quad
\widehat{\operatorname{df}}_n(\lambda_1)
>
\widehat{\operatorname{df}}_n(\lambda_2)
\]
for all $\lambda_1,\lambda_2>-\hat\mu_m$.
\end{proposition}

Finally, the anti-shrinkage effect can be written directly as a deviation from the restricted oracle.

\begin{proposition}[Distance from the restricted oracle]
\label{prop:distance_from_restricted_oracle}
For every $\lambda>-\mu_m$,
\[
\beta_\lambda^{\mathrm{pop}}-\beta^\dagger
=
-\lambda(\Sigma+\lambda I_m)^{-1}\beta^\dagger,
\]
and therefore
\[
\|\beta_\lambda^{\mathrm{pop}}-\beta^\dagger\|_2^2
=
\sum_{j=1}^m
\frac{\lambda^2}{(\mu_j+\lambda)^2}\,\alpha_j^2.
\]
\end{proposition}

Together, Proposition~\ref{prop:spectral_form_population}--Proposition~\ref{prop:distance_from_restricted_oracle} show that negative regularization increases complexity through a structured spectral filter rather than through indiscriminate destabilization. Full proofs are deferred to Appendix~\ref{app:proofs_feasible_complexity}.

\subsection{Weak-Spectrum Underfitting}
\label{subsec:theory_weak_spectrum}

The previous subsection established that negative regularization expands the estimator most strongly along eigendirections with small covariance eigenvalues. To turn this observation into a theory of underfitting, we now formalize the structural regime in which the restricted oracle itself is substantially aligned with those weak directions. This is the regime in which ordinary nonnegative shrinkage can be especially damaging, because it suppresses precisely the coordinates that remain necessary for accurate prediction within the restricted representation space.

Recall the eigendecomposition
\[
\Sigma = U \operatorname{diag}(\mu_1,\dots,\mu_m) U^\top,
\qquad
\mu_1 \ge \cdots \ge \mu_m > 0,
\]
and the expansion
\[
\beta^\dagger = \sum_{j=1}^m \alpha_j u_j.
\]
The quantity $\alpha_j$ measures how much of the restricted oracle lies in the $j$th eigendirection, while $\mu_j$ measures how strongly that direction is supported by the representation covariance. Weak-spectrum underfitting occurs when a nontrivial part of the signal geometry is carried by indices with comparatively small $\mu_j$.

To formalize this, fix a threshold $\kappa \in [\mu_m,\mu_1]$ and define the weak and strong index sets by
\[
W_\kappa := \{j \in \{1,\dots,m\} : \mu_j \le \kappa\},
\qquad
S_\kappa := \{j \in \{1,\dots,m\} : \mu_j > \kappa\}.
\]
These induce the orthogonal decomposition
\[
\beta^\dagger
=
\beta^\dagger_{W,\kappa} + \beta^\dagger_{S,\kappa},
\]
where
\[
\beta^\dagger_{W,\kappa}
:=
\sum_{j \in W_\kappa} \alpha_j u_j,
\qquad
\beta^\dagger_{S,\kappa}
:=
\sum_{j \in S_\kappa} \alpha_j u_j.
\]

\begin{definition}[Weak-spectrum signal masses]
\label{def:weak_spectrum_masses}
For a threshold $\kappa \in [\mu_m,\mu_1]$, define
\[
A_W(\kappa)
:=
\|\beta^\dagger_{W,\kappa}\|_2^2
=
\sum_{j \in W_\kappa} \alpha_j^2,
\qquad
A_S(\kappa)
:=
\|\beta^\dagger_{S,\kappa}\|_2^2
=
\sum_{j \in S_\kappa} \alpha_j^2,
\]
and
\[
M_W(\kappa)
:=
\|\Sigma^{1/2}\beta^\dagger_{W,\kappa}\|_2^2
=
\sum_{j \in W_\kappa} \mu_j \alpha_j^2,
\qquad
M_S(\kappa)
:=
\|\Sigma^{1/2}\beta^\dagger_{S,\kappa}\|_2^2
=
\sum_{j \in S_\kappa} \mu_j \alpha_j^2.
\]
We also define the weak-spectrum alignment ratio
\[
\rho_W(\kappa)
:=
\frac{A_W(\kappa)}{A_W(\kappa)+A_S(\kappa)}
\]
whenever $\beta^\dagger \neq 0$.
\end{definition}

Here, $A_W(\kappa)$ measures how much Euclidean mass of the restricted oracle lies in weak directions, while $M_W(\kappa)$ measures how much prediction-weighted mass lies there. The distinction matters because a direction may carry a large coefficient while still lying in a weak eigendirection of the representation.

\begin{proposition}[Spectral sandwich on weak and strong subspaces]
\label{prop:spectral_sandwich_weak_strong}
For every $\kappa \in [\mu_m,\mu_1]$,
\[
\mu_m A_W(\kappa)
\le
M_W(\kappa)
\le
\kappa A_W(\kappa),
\]
and
\[
\kappa A_S(\kappa)
<
M_S(\kappa)
\le
\mu_1 A_S(\kappa).
\]
Equivalently,
\[
\|\Sigma^{1/2}\beta^\dagger_{W,\kappa}\|_2^2
\le
\kappa \|\beta^\dagger_{W,\kappa}\|_2^2,
\qquad
\|\Sigma^{1/2}\beta^\dagger_{S,\kappa}\|_2^2
>
\kappa \|\beta^\dagger_{S,\kappa}\|_2^2.
\]
\end{proposition}

The proposition makes the central asymmetry explicit: a given amount of Euclidean oracle mass costs less in prediction norm when placed in weak directions than in strong ones. This is precisely why shrinkage can create substantial coefficient distortion before it becomes equally visible in prediction norm.

To connect this structure to regularization bias, define the population excess risk of the regularized target by
\[
B(\lambda)
:=
\mathcal{E}\!\left(\beta_\lambda^{\mathrm{pop}}\right)
=
\mathcal{R}\!\left(\beta_\lambda^{\mathrm{pop}}\right)-\mathcal{R}(\beta^\dagger),
\qquad
\lambda>-\mu_m.
\]

\begin{proposition}[Bias decomposition across weak and strong directions]
\label{prop:bias_decomposition_weak_strong}
For every $\lambda > -\mu_m$ and every threshold $\kappa \in [\mu_m,\mu_1]$,
\[
B(\lambda)
=
B_W(\lambda;\kappa)+B_S(\lambda;\kappa),
\]
where
\[
B_W(\lambda;\kappa)
:=
\sum_{j \in W_\kappa}
\mu_j
\frac{\lambda^2}{(\mu_j+\lambda)^2}\alpha_j^2,
\qquad
B_S(\lambda;\kappa)
:=
\sum_{j \in S_\kappa}
\mu_j
\frac{\lambda^2}{(\mu_j+\lambda)^2}\alpha_j^2.
\]
Moreover, for each fixed $\lambda \in (-\mu_m,\infty)\setminus\{0\}$, the directional bias multiplier
\[
b_\lambda(\mu)
:=
\mu \frac{\lambda^2}{(\mu+\lambda)^2}
\]
satisfies
\[
b_\lambda'(\mu)
=
\frac{\lambda^2(\lambda-\mu)}{(\mu+\lambda)^3}.
\]
In particular, if $\lambda<0$, then $b_\lambda'(\mu)<0$ for all $\mu>-\lambda$.
\end{proposition}

Thus, once $\lambda$ enters the negative region, the regularization-induced bias becomes spectrally ordered: smaller eigenvalues contribute more strongly to the directional bias multiplier.

\begin{proposition}[Weak directions are the most shrinkage-sensitive]
\label{prop:weak_directions_most_sensitive}
Fix $\lambda \in (-\mu_m,0)$ and let
\[
b_j(\lambda)
:=
\mu_j\frac{\lambda^2}{(\mu_j+\lambda)^2}.
\]
If $\mu_i \ge \mu_j$, then
\[
b_i(\lambda)\le b_j(\lambda),
\]
with strict inequality whenever $\mu_i>\mu_j$.
Consequently, for every threshold $\kappa \in [\mu_m,\mu_1]$,
\[
B_W(\lambda;\kappa)
\ge
\min_{j \in W_\kappa} b_j(\lambda)\, A_W(\kappa),
\qquad
B_S(\lambda;\kappa)
\le
\max_{j \in S_\kappa} b_j(\lambda)\, A_S(\kappa),
\]
and
\[
\min_{j \in W_\kappa} b_j(\lambda)
\ge
\max_{j \in S_\kappa} b_j(\lambda)
\]
whenever both index sets are nonempty.
\end{proposition}

These bounds motivate the structural regime of interest.

\begin{definition}[Weak-spectrum underfitting]
\label{def:weak_spectrum_underfitting}
We say that the restricted regression problem exhibits \emph{weak-spectrum underfitting} at threshold $\kappa \in [\mu_m,\mu_1]$ if
\[
A_W(\kappa) > 0,
\]
and a nonnegligible fraction of the restricted oracle mass is concentrated in $W_\kappa$, i.e.,
\[
\rho_W(\kappa)
=
\frac{A_W(\kappa)}{A_W(\kappa)+A_S(\kappa)}
\]
is bounded away from zero.
A stronger form holds when
\[
A_W(\kappa)\ge c_0 \|\beta^\dagger\|_2^2
\]
for some constant $c_0 \in (0,1]$.
\end{definition}

This definition is intentionally structural rather than algorithmic. It does not say that the observed estimator already underfits numerically; rather, it identifies a geometry in which underfitting is likely because the coordinates that matter inside the restricted model are precisely those that ordinary shrinkage suppresses most strongly.

\begin{proposition}[Bias lower bound under weak-spectrum alignment]
\label{prop:bias_lower_bound_weak_alignment}
Fix $\kappa \in [\mu_m,\mu_1]$ and $\lambda \in (-\mu_m,0)$. Then
\[
B(\lambda)
\ge
B_W(\lambda;\kappa)
\ge
\left(
\min_{j \in W_\kappa}
\mu_j\frac{\lambda^2}{(\mu_j+\lambda)^2}
\right)
A_W(\kappa).
\]
In particular, if the problem satisfies the strong weak-spectrum underfitting condition
\[
A_W(\kappa)\ge c_0 \|\beta^\dagger\|_2^2,
\]
then
\[
B(\lambda)
\ge
c_0
\left(
\min_{j \in W_\kappa}
\mu_j\frac{\lambda^2}{(\mu_j+\lambda)^2}
\right)
\|\beta^\dagger\|_2^2.
\]
\end{proposition}

Proposition~\ref{prop:bias_lower_bound_weak_alignment} makes explicit that weak-spectrum alignment alone already forces a nontrivial lower bound on regularization bias in the negative-capable family. Proofs for the results in this subsection are deferred to Appendix~\ref{app:proofs_weak_spectrum_signswitch}.

\subsection{Sign-Switch Theorem}
\label{subsec:theory_sign_switch}

A genuine sign-switch cannot be established from the population target
\[
\beta_\lambda^{\mathrm{pop}}=(\Sigma+\lambda I_m)^{-1}\Sigma\beta^\dagger
\]
alone, because its excess risk relative to the restricted oracle is minimized at the unregularized point $\lambda=0$. The sign-switch question therefore has to be posed relative to a \emph{baseline conservative regime}: the small-data learner is already operating under a positive amount of effective shrinkage, and the question is whether the oracle adjustment away from that baseline should move into the negative region.

To formalize this, fix a baseline conservativeness level $\tau_0>0$ and write
\[
\eta(\lambda):=\tau_0+\lambda.
\]
We study local adjustments $\lambda$ around the baseline $\tau_0$, with feasible range
\[
\lambda>-\tau_0-\mu_m,
\]
so that the net shrinkage level $\eta(\lambda)$ remains above the spectral boundary $-\mu_m$. The negative-adjustment region is therefore
\[
\Lambda_{-}(\tau_0):=(-\tau_0,0).
\]

We model the oracle linearized estimator by
\[
\tilde\beta_\lambda
:=
(\Sigma+\eta(\lambda)I_m)^{-1}\bigl(\Sigma\beta^\dagger+\xi_n\bigr),
\]
where $\xi_n\in\mathbb{R}^m$ is a mean-zero noise term satisfying
\[
\mathbb{E}[\xi_n]=0,
\qquad
\operatorname{Cov}(\xi_n)=\frac{\sigma^2}{n}\Sigma.
\]
This deterministic-covariance surrogate isolates the two effects that matter for the sign-switch analysis: decreasing $\lambda$ reduces the shrinkage bias induced by the baseline level $\tau_0$, but it also inflates estimation variance.

We measure performance by the expected excess prediction risk
\[
\mathfrak{R}_n(\lambda;\tau_0)
:=
\mathbb{E}\bigl[\mathcal{E}(\tilde\beta_\lambda)\bigr].
\]

\begin{proposition}[Oracle criterion under a conservative baseline]
\label{prop:oracle_criterion_baseline}
For every $\lambda>-\tau_0-\mu_m$,
\[
\mathfrak{R}_n(\lambda;\tau_0)
=
\sum_{j=1}^m
\mu_j
\frac{\eta(\lambda)^2\alpha_j^2}{(\mu_j+\eta(\lambda))^2}
+
\frac{\sigma^2}{n}
\sum_{j=1}^m
\frac{\mu_j^2}{(\mu_j+\eta(\lambda))^2}.
\]
Equivalently,
\[
\mathfrak{R}_n(\lambda;\tau_0)
=
\mathfrak{B}_n(\lambda;\tau_0)
+
\mathfrak{V}_n(\lambda;\tau_0),
\]
where
\[
\mathfrak{B}_n(\lambda;\tau_0)
:=
\sum_{j=1}^m
\mu_j
\frac{\eta(\lambda)^2\alpha_j^2}{(\mu_j+\eta(\lambda))^2}
\]
is the baseline-shrinkage bias term and
\[
\mathfrak{V}_n(\lambda;\tau_0)
:=
\frac{\sigma^2}{n}
\sum_{j=1}^m
\frac{\mu_j^2}{(\mu_j+\eta(\lambda))^2}
\]
is the variance term.
\end{proposition}

This decomposition makes the role of the baseline explicit: moving leftward reduces the first term by undoing part of the baseline shrinkage, but increases the second term through variance inflation.

\begin{proposition}[Derivative of the oracle criterion]
\label{prop:derivative_oracle_criterion}
For every $\lambda>-\tau_0-\mu_m$,
\[
\frac{\partial}{\partial \lambda}
\mathfrak{R}_n(\lambda;\tau_0)
=
2
\sum_{j=1}^m
\frac{\mu_j^2}{(\mu_j+\eta(\lambda))^3}
\left(
\eta(\lambda)\alpha_j^2-\frac{\sigma^2}{n}
\right).
\]
In particular, at the baseline point $\lambda=0$,
\[
\frac{\partial}{\partial \lambda}
\mathfrak{R}_n(0;\tau_0)
=
2
\sum_{j=1}^m
\frac{\mu_j^2}{(\mu_j+\tau_0)^3}
\left(
\tau_0\alpha_j^2-\frac{\sigma^2}{n}
\right).
\]
\end{proposition}

The derivative formula isolates the sign-switch mechanism sharply: a negative adjustment is locally preferred when the bias relief created by reducing baseline shrinkage dominates the corresponding variance penalty.

\begin{theorem}[Local sign-switch theorem]
\label{thm:local_sign_switch}
Assume that
\[
\frac{\partial}{\partial \lambda}
\mathfrak{R}_n(0;\tau_0)>0,
\]
or equivalently,
\[
\tau_0
\sum_{j=1}^m
\frac{\mu_j^2\alpha_j^2}{(\mu_j+\tau_0)^3}
>
\frac{\sigma^2}{n}
\sum_{j=1}^m
\frac{\mu_j^2}{(\mu_j+\tau_0)^3}.
\]
Then there exists $\delta\in(0,\tau_0)$ such that
\[
\mathfrak{R}_n(-\delta;\tau_0)
<
\mathfrak{R}_n(0;\tau_0).
\]
Consequently, if
\[
\lambda_n^\star(\tau_0)
\in
\arg\min_{\lambda>-\tau_0-\mu_m}\mathfrak{R}_n(\lambda;\tau_0),
\]
then at least one oracle minimizer satisfies
\[
\lambda_n^\star(\tau_0)<0.
\]
\end{theorem}

The theorem shows that the sign-switch is inherently relative to an already conservative baseline; it does not claim that negative regularization is universally preferable to zero or positive regularization.

We now convert this derivative condition into a structural sufficient condition expressed in terms of the weak-spectrum quantities introduced in the previous subsection.

\begin{proposition}[Weak-spectrum sufficient condition for a positive derivative]
\label{prop:weak_spectrum_sufficient_condition}
Fix a threshold $\kappa\in[\mu_m,\mu_1]$ and define
\[
w_j(\tau_0):=\frac{\mu_j^2}{(\mu_j+\tau_0)^3}.
\]
Then
\[
\frac{\partial}{\partial \lambda}
\mathfrak{R}_n(0;\tau_0)
\ge
2\tau_0
\left(
\min_{j\in W_\kappa} w_j(\tau_0)
\right)
A_W(\kappa)
-
\frac{2\sigma^2}{n}
\sum_{j=1}^m
w_j(\tau_0).
\]
In particular, if
\[
\tau_0
\left(
\min_{j\in W_\kappa} w_j(\tau_0)
\right)
A_W(\kappa)
>
\frac{\sigma^2}{n}
\sum_{j=1}^m
w_j(\tau_0),
\]
then the condition of Theorem~\ref{thm:local_sign_switch} holds.
\end{proposition}

\begin{corollary}[Sign-switch under strong weak-spectrum underfitting]
\label{cor:sign_switch_strong_weak_spectrum}
Suppose that, for some $\kappa\in[\mu_m,\mu_1]$ and some $c_0\in(0,1]$,
\[
A_W(\kappa)\ge c_0\|\beta^\dagger\|_2^2.
\]
If
\[
\tau_0 c_0
\left(
\min_{j\in W_\kappa} \frac{\mu_j^2}{(\mu_j+\tau_0)^3}
\right)
\|\beta^\dagger\|_2^2
>
\frac{\sigma^2}{n}
\sum_{j=1}^m
\frac{\mu_j^2}{(\mu_j+\tau_0)^3},
\]
then there exists $\delta\in(0,\tau_0)$ such that
\[
\mathfrak{R}_n(-\delta;\tau_0)
<
\mathfrak{R}_n(0;\tau_0),
\]
and hence at least one oracle adjustment satisfies
\[
\lambda_n^\star(\tau_0)<0.
\]
\end{corollary}

These results show that the sign-switch is not a generic property of negative adjustment. It arises when conservative baseline shrinkage interacts with sufficiently strong weak-spectrum alignment and sufficiently low noise. Full proofs are deferred to Appendix~\ref{app:proofs_weak_spectrum_signswitch}.

\subsection{Automatic Selection}
\label{subsec:theory_selection}

We now turn from oracle tuning to data-driven selection over the negative-capable family. Since the sign-switch theorem was formulated relative to a baseline conservative level $\tau_0>0$, we keep the same parameterization and write
\[
\eta(\lambda):=\tau_0+\lambda.
\]
Fix constants $\tau\in(0,\tau_0)$ and $L>0$, and let
\[
\Gamma_n \subset [-\tau_0+\tau,\,L]
\]
be a finite candidate set. The lower truncation by $\tau$ ensures that
\[
\eta(\lambda)\ge \tau>0
\qquad
\text{for all } \lambda\in\Gamma_n,
\]
so every candidate remains in the interior of the feasible region.

For each $\lambda\in\Gamma_n$, define the baseline-adjusted ridge estimator
\[
\hat\beta_\lambda^{(\tau_0)}
:=
\bigl(\hat\Sigma_n+\eta(\lambda)I_m\bigr)^{-1}\hat g_n,
\]
and the corresponding fitted mean vector
\[
\hat m_\lambda
:=
\Phi \hat\beta_\lambda^{(\tau_0)}
=
H_\lambda Y,
\]
where
\[
H_\lambda
:=
\Phi\bigl(\Phi^\top\Phi+n\eta(\lambda)I_m\bigr)^{-1}\Phi^\top
=
\frac{1}{n}\Phi\bigl(\hat\Sigma_n+\eta(\lambda)I_m\bigr)^{-1}\Phi^\top.
\]
Because $\eta(\lambda)>0$, the matrix $H_\lambda$ is symmetric.

Let
\[
m_n
:=
\begin{bmatrix}
f_\star(x_1)\\
\vdots\\
f_\star(x_n)
\end{bmatrix}
\in\mathbb{R}^n,
\qquad
Y=m_n+\varepsilon,
\]
where, conditional on the design $\Phi$, we assume
\[
\mathbb{E}[\varepsilon\mid \Phi]=0,
\qquad
\mathbb{E}[\varepsilon\varepsilon^\top\mid \Phi]=\sigma^2 I_n.
\]
We measure performance by the conditional prediction risk
\[
\mathcal{P}_n(\lambda)
:=
\frac{1}{n}\,
\mathbb{E}\bigl[\|m_n-\hat m_\lambda\|_2^2 \mid \Phi\bigr]
=
\frac{1}{n}\,
\mathbb{E}\bigl[\|m_n-H_\lambda Y\|_2^2 \mid \Phi\bigr].
\]

\begin{proposition}[Conditional unbiased risk estimate]
\label{prop:conditional_unbiased_risk_estimate}
Assume that $\sigma^2$ is known. For each $\lambda\in\Gamma_n$, define
\[
\operatorname{Crit}_n(\lambda)
:=
\frac{1}{n}\|Y-\hat m_\lambda\|_2^2
+
\frac{2\sigma^2}{n}\operatorname{tr}(H_\lambda)
-
\sigma^2.
\]
Then
\[
\mathbb{E}\bigl[\operatorname{Crit}_n(\lambda)\mid \Phi\bigr]
=
\mathcal{P}_n(\lambda)
\qquad
\text{for all } \lambda\in\Gamma_n.
\]
\end{proposition}

Thus $\operatorname{Crit}_n(\lambda)$ is a conditionally unbiased SURE-type criterion over the full negative-capable family, not merely over the nonnegative subfamily.

We now define the data-driven selector
\[
\hat\lambda_n
\in
\arg\min_{\lambda\in\Gamma_n}\operatorname{Crit}_n(\lambda).
\]

\begin{theorem}[Oracle inequality for criterion-based selection]
\label{thm:oracle_inequality_selection}
Define the uniform criterion error
\[
\Delta_n
:=
\sup_{\lambda\in\Gamma_n}
\bigl|
\operatorname{Crit}_n(\lambda)-\mathcal{P}_n(\lambda)
\bigr|.
\]
Then, conditional on $\Phi$,
\[
\mathcal{P}_n(\hat\lambda_n)
\le
\inf_{\lambda\in\Gamma_n}\mathcal{P}_n(\lambda)
+
2\Delta_n.
\]
Consequently,
\[
\mathbb{E}\bigl[\mathcal{P}_n(\hat\lambda_n)\mid \Phi\bigr]
\le
\inf_{\lambda\in\Gamma_n}\mathcal{P}_n(\lambda)
+
2\mathbb{E}[\Delta_n\mid \Phi].
\]
\end{theorem}

The theorem shows that the selected adjustment is as good as the best candidate in the discrete family up to twice the uniform criterion-estimation error.

Define
\[
\Gamma_n^-:=\Gamma_n\cap[-\tau_0+\tau,\,0),
\qquad
\Gamma_n^+:=\Gamma_n\cap[0,\,L].
\]
Assume both sets are nonempty.

\begin{proposition}[Recovery of a negative adjustment under a margin condition]
\label{prop:negative_recovery_margin}
Suppose
\[
\inf_{\lambda\in\Gamma_n^+}\mathcal{P}_n(\lambda)
-
\inf_{\lambda\in\Gamma_n^-}\mathcal{P}_n(\lambda)
>
2\Delta_n.
\]
Then
\[
\hat\lambda_n\in\Gamma_n^-.
\]
\end{proposition}

This proposition connects selection back to the sign-switch theory: if the negative part of the candidate family is genuinely better than the nonnegative part by a margin exceeding the uniform criterion error, the selector must recover a negative adjustment.

Finally, we record a simple approximation statement for the passage from a continuous negative-capable interval to a discrete candidate grid.

Let
\[
\Lambda_n^{\mathrm{cont}}
:=
[-\tau_0+\tau,\,L]
\]
and suppose that the finite grid $\Gamma_n\subset \Lambda_n^{\mathrm{cont}}$ has mesh width
\[
h_n
:=
\sup_{\lambda\in \Lambda_n^{\mathrm{cont}}}
\min_{\tilde\lambda\in\Gamma_n} |\lambda-\tilde\lambda|.
\]

\begin{proposition}[Discrete approximation of the continuous oracle]
\label{prop:grid_approximation}
Assume that there exists a constant $C_n<\infty$ such that
\[
|\mathcal{P}_n(\lambda_1)-\mathcal{P}_n(\lambda_2)|
\le
C_n |\lambda_1-\lambda_2|
\qquad
\text{for all } \lambda_1,\lambda_2\in\Lambda_n^{\mathrm{cont}}.
\]
Then
\[
\inf_{\lambda\in\Gamma_n}\mathcal{P}_n(\lambda)
\le
\inf_{\lambda\in\Lambda_n^{\mathrm{cont}}}\mathcal{P}_n(\lambda)
+
C_n h_n.
\]
Consequently,
\[
\mathcal{P}_n(\hat\lambda_n)
\le
\inf_{\lambda\in\Lambda_n^{\mathrm{cont}}}\mathcal{P}_n(\lambda)
+
C_n h_n
+
2\Delta_n.
\]
\end{proposition}

Taken together, Proposition~\ref{prop:conditional_unbiased_risk_estimate}--Proposition~\ref{prop:grid_approximation} justify automatic tuning over the full negative-capable family. The criterion is conditionally unbiased, the selected parameter enjoys an oracle inequality, a sufficiently favorable negative-region margin forces negative selection, and a fine enough grid tracks the continuous oracle up to an explicit discretization term. Detailed proofs are deferred to Appendix~\ref{app:proofs_selection}.

\section{Experiments}
\label{sec:experiments}

\subsection{Verification of the Feasible Negative Interval and Effective Complexity Increase}
\label{subsec:exp_feasible_complexity}

\paragraph{Experimental setup.}
Our experiments were designed to evaluate the proposed negative-capable regularization family from four complementary angles: feasibility, structural mechanism, oracle sign-switch, and automatic selection. The first experiment, reported in Table~\ref{tab:exp1_feasible_complexity}, examines whether a nontrivial negative region is empirically well-posed and how moving the regularization parameter across negative, zero, and positive values changes effective complexity and predictive behavior. To this end, we consider a small-data synthetic regression setting with a restricted representation and a weak-spectrum-aligned signal, and we evaluate representative values of the regularization parameter spanning the feasible negative interval, the unregularized point, and the positive region. For each value, we record its distance from the empirical spectral boundary, the empirical effective degrees of freedom, the coefficient norm, and both training and test mean squared error.

\paragraph{Results and analysis.}
Table~\ref{tab:exp1_feasible_complexity} shows that the empirical feasible negative interval is nonempty in this setting, so negative regularization is admissible for a nontrivial range of parameter values. As the regularization parameter moves leftward, the empirical effective degrees of freedom and coefficient magnitude increase monotonically. The table also shows that performance worsens when the parameter approaches the empirical spectral boundary too closely, even though all reported values remain within the feasible range.

\begin{table}[ht]
\centering
\caption{Feasibility of the negative region and its effect on effective complexity.}
\label{tab:exp1_feasible_complexity}
\begin{tabular}{rcrrrrr}
\toprule
$\lambda$ & Sign & Spectral margin & Empirical df & $\|\hat{\beta}_{\lambda}\|_2$ & Train MSE & Test MSE \\
\midrule
-0.0180 & \multirow{3}{*}{Negative} & 0.0032 & 18.8573 & 3.0664 & 0.2836 & 0.4768 \\
-0.0127 &                           & 0.0085 & 14.2674 & 1.4238 & 0.1617 & 0.1736 \\
-0.0053 &                           & 0.0159 & 12.6206 & 1.0331 & 0.1521 & 0.1521 \\
\midrule
\hphantom{-}0.0000 & Zero           & 0.0212 & 12.0000 & 0.9263 & 0.1515 & 0.1499 \\
\midrule
\hphantom{-}0.0100 & \multirow{3}{*}{Positive} & 0.0312 & 11.2227 & 0.8148 & 0.1523 & 0.1488 \\
\hphantom{-}0.0500 &                           & 0.0712 & 9.6162  & 0.6178 & 0.1597 & 0.1477 \\
\hphantom{-}0.2000 &                           & 0.2212 & 7.3148  & 0.3720 & 0.1847 & 0.1545 \\
\bottomrule
\end{tabular}
\end{table}

\subsection{Verification of Weak-Spectrum Underfitting Structure}
\label{subsec:exp_weak_spectrum}

\paragraph{Experimental setup.}
The second experiment examines the structural pattern of weak-spectrum underfitting by comparing three oracle alignments, namely \emph{weak}, \emph{balanced}, and \emph{strong}. In all cases, the covariance spectrum is fixed, and a common negative regularization level is used so that differences arise only from how the restricted oracle is distributed across weak and strong eigendirections. Table~\ref{tab:exp2_weak_spectrum} reports the fraction of oracle mass lying in the weak spectrum, the weak and strong components of the coefficient mass, their covariance-weighted counterparts, and the corresponding decomposition of the regularization-induced bias. This design isolates the spectral geometry of underfitting rather than predictive performance itself.

\paragraph{Results and analysis.}
Table~\ref{tab:exp2_weak_spectrum} shows a clear progression across the three alignment settings. Under weak alignment, most of the oracle mass lies in the weak spectrum, and the regularization-induced bias is correspondingly concentrated there. Under balanced alignment, the weak component of the oracle is smaller, but the bias remains strongly tilted toward the weak directions. Under strong alignment, both the weak-spectrum mass and its bias contribution become much smaller relative to the strong component. These results verify that the weak-spectrum structure is determined by how strongly the restricted oracle aligns with low-eigenvalue directions.

\begin{table}[ht]
\centering
\caption{Weak-spectrum underfitting structure across oracle alignments.}
\label{tab:exp2_weak_spectrum}
\begin{tabular}{lrrrrrrrr}
\toprule
Alignment & $\rho_W(\kappa)$ & $A_W(\kappa)$ & $A_S(\kappa)$ & $M_W(\kappa)$ & $M_S(\kappa)$ & $B_W(\lambda;\kappa)$ & $B_S(\lambda;\kappa)$ & Weak bias share \\
\midrule
Weak     & 0.9198 & 0.6720 & 0.0586 & 0.0527 & 0.0718 & 0.0139 & 0.0001 & 0.9959 \\
Balanced & 0.2175 & 0.0400 & 0.1439 & 0.0040 & 0.3114 & 0.0006 & 0.0001 & 0.9151 \\
Strong   & 0.0074 & 0.0054 & 0.7288 & 0.0007 & 2.1115 & 0.0000 & 0.0001 & 0.2387 \\
\bottomrule
\end{tabular}
\end{table}

\subsection{Sign-Switch under Baseline Conservativeness}
\label{subsec:exp_sign_switch}

\paragraph{Experimental setup.}
The third experiment evaluates the sign-switch phenomenon under a conservative baseline by comparing multiple synthetic scenarios that vary oracle alignment, noise level, and baseline shrinkage. We consider \emph{weak}, \emph{balanced}, and \emph{strong} oracle alignments, and for each setting we specify a noise scale and a baseline regularization level. For every scenario, we compute the derivative of the oracle risk with respect to the adjustment parameter at the baseline point, record whether this local criterion predicts a negative adjustment, and then compare it with the sign of the oracle adjustment obtained by direct risk minimization over the admissible adjustment range. Table~\ref{tab:exp3_sign_switch} summarizes these quantities.

\paragraph{Results and analysis.}
Table~\ref{tab:exp3_sign_switch} shows that the local derivative criterion and the oracle adjustment sign agree across all tested scenarios. Negative oracle adjustments appear consistently under weak alignment, do not appear under balanced alignment, and arise only conditionally under strong alignment depending on the noise level. The results therefore verify that the sign-switch pattern is tied to the joint effect of oracle alignment, noise, and baseline conservativeness, rather than being a universal property of negative adjustment.

\begin{table}[ht]
\centering
\caption{Sign-switch under baseline conservativeness.}
\label{tab:exp3_sign_switch}
\begin{tabular}{lrrrrrl}
\toprule
Align. & $\sigma$ & $\tau_0$ & $\partial R(0)$ & Local neg. & $\lambda^\star$ & Sign \\
\midrule
\multirow{2}{*}{Weak} & 0.1500 & 0.2000 &  0.0798 & Yes & -0.1960 & Negative \\
                      & 0.5000 & 0.2000 &  0.0290 & Yes & -0.1480 & Negative \\
\midrule
\multirow{2}{*}{Balanced} & 0.3000 & 0.1000 & -0.0130 & No  &  0.0439 & Positive \\
                          & 0.5000 & 0.2000 & -0.0261 & No  &  0.1367 & Positive \\
\midrule
\multirow{2}{*}{Strong} & 0.3000 & 0.1000 &  0.0165 & Yes & -0.0193 & Negative \\
                        & 0.5000 & 0.1000 & -0.0515 & No  &  0.0494 & Positive \\
\bottomrule
\end{tabular}
\end{table}

\subsection{Automatic Selection over the Negative-Capable Family}
\label{subsec:exp_selection}

\paragraph{Experimental setup.}
The fourth experiment evaluates whether a data-driven selector can effectively operate over the full negative-capable family rather than over a nonnegative subfamily only. We consider four controlled synthetic scenarios spanning different oracle alignments, noise levels, and baseline shrinkage values, and repeat each scenario over multiple independently generated training--test splits. For each repetition, we compare three quantities: an empirical oracle obtained by directly minimizing test error over the full adjustment grid, a \emph{positive-only} selector that searches only over nonnegative adjustments, and a \emph{negative-capable} selector that searches over the full adjustment grid including negative values. The selector itself is based on a SURE-type criterion, and Table~\ref{tab:exp4_selection} reports the frequency with which the oracle and the selector choose negative adjustments, together with the average test errors of the positive-only selector, the negative-capable selector, and the empirical oracle, as well as the corresponding performance gaps.

\paragraph{Results and analysis.}
Table~\ref{tab:exp4_selection} shows that the negative-capable selector is most effective in the regime where negative adjustment is most consistently favored by the oracle. In that setting, the selector tracks the oracle closely both in sign choice and in predictive performance, and the negative-capable family improves substantially over the positive-only alternative. In the remaining settings, the oracle selects negative adjustments less consistently, and the empirical advantage of the negative-capable selector becomes correspondingly weaker. The table therefore shows a clear dependence of automatic selection performance on the structural strength of the sign-switch regime: when negative adjustment is strongly supported, the selector recovers that advantage well, whereas in more marginal settings the gains become small and the selector behaves more conservatively. 

\begin{table}[ht]
\centering
\caption{Automatic selection over the negative-capable family.}
\label{tab:exp4_selection}
\begin{tabular}{lrrrrrrrrr}
\toprule
Align. & $\sigma$ & $\tau_0$ & O-neg & S-neg & MSE$+$ & MSE$\pm$ & MSE$^\star$ & $\Delta_{\pm,+}$ & Gap \\
\midrule
Weak (low noise)  & 0.1500 & 0.2000 & 1.0000 & 1.0000 & 0.0603 & 0.0305 & 0.0300 & -0.0299 & 0.0004 \\
Weak (high noise) & 0.5000 & 0.2000 & 0.6250 & 0.6500 & 0.3252 & 0.3295 & 0.3091 &  0.0043 & 0.0204 \\
Balanced & 0.3000 & 0.1000 & 0.4125 & 0.2375 & 0.1119 & 0.1129 & 0.1091 &  0.0010 & 0.0038 \\
Strong   & 0.5000 & 0.1000 & 0.3125 & 0.1000 & 0.2957 & 0.2969 & 0.2893 &  0.0012 & 0.0076 \\
\bottomrule
\end{tabular}
\end{table}

\subsection{Real-Data Small-Data Regression under Representation Bottlenecks}
\label{subsec:exp_real_data}

\paragraph{Experimental setup.}
The fifth experiment evaluates the proposed method on a semi-synthetic benchmark built on top of real regression datasets. For each dataset, we first construct a fixed low-dimensional representation through unsupervised preprocessing and PCA-based compression, thereby imposing an explicit representation bottleneck while preserving the empirical covariance structure of the original inputs. Synthetic responses are then generated on this real covariance geometry under three oracle alignment regimes---\emph{weak}, \emph{balanced}, and \emph{strong}---and under two noise conditions, \emph{low} and \emph{high}. Small-data conditions are created by repeated subsampling of the training pool at several training fractions. Within each setting, we compare a \emph{positive-only} selector with a \emph{negative-capable} selector over the baseline-adjusted ridge family. Table~\ref{tab:exp5_main_summary} reports the main summary by alignment and noise, and Table~\ref{tab:exp5_trainfrac_summary} reports the complementary summary by alignment and training fraction.

\paragraph{Results and analysis.}
Table~\ref{tab:exp5_main_summary} shows that negative adjustment is strongly supported across all alignment and noise settings: the oracle selects negative adjustments at very high rates, and the data-driven selector closely tracks that behavior. In every reported setting, the negative-capable selector achieves lower prediction error than the positive-only selector, with larger gains appearing when the weak-spectrum component is stronger and when the noise level is lower. The gap to the oracle remains small, indicating that the empirical gain is not merely an oracle phenomenon but is recovered effectively by automatic tuning. Table~\ref{tab:exp5_trainfrac_summary} further shows that this pattern is stable across all training fractions considered here. The negative-capable selector continues to outperform the positive-only alternative throughout the small-data regime, while its gap to the oracle decreases as the training fraction increases.

\begin{table}[ht]
\centering
\caption{Main summary by alignment and noise.}
\label{tab:exp5_main_summary}
\begin{tabular}{llrrrrrrrr}
\toprule
Align. & Noise & $\rho_W$ & O-neg & S-neg & RMSE$+$ & RMSE$\pm$ & Gain (\%) & Gap & Win \\
\midrule
\multirow{2}{*}{Weak} & High & 0.8667 & 1.0000 & 1.0000 & 0.7328 & 0.5795 & 20.9150 & 0.0067 & 0.9762 \\
                      & Low  & 0.8667 & 1.0000 & 1.0000 & 0.5608 & 0.2786 & 50.3172 & 0.0000 & 1.0000 \\
\midrule
\multirow{2}{*}{Balanced} & High & 0.4506 & 0.9810 & 0.9905 & 0.6367 & 0.5587 & 12.2587 & 0.0086 &  0.9333 \\
                          & Low  & 0.4506 & 1.0000 & 1.0000 & 0.4194 & 0.2443 & 41.7647 & 0.0008 & 0.9976 \\
\midrule
\multirow{2}{*}{Strong}   & High & 0.0826 & 0.9524 & 0.9500 & 0.5846 & 0.5410 & 7.4624  & 0.0092 & 0.8190 \\
                          & Low  & 0.0826 & 1.0000 & 1.0000 & 0.3407 & 0.2276 & 33.2033 & 0.0032 & 0.9786 \\
\bottomrule
\end{tabular}
\end{table}

\begin{table}[ht]
\centering
\caption{Summary by alignment and training fraction.}
\label{tab:exp5_trainfrac_summary}
\begin{tabular}{lrrrrrrrr}
\toprule
Align. & Frac. & O-neg & S-neg & RMSE$+$ & RMSE$\pm$ & Gain (\%) & Gap & Win \\
\midrule
\multirow{3}{*}{Weak} & 0.0200 & 1.0000 & 1.0000 & 0.6614 & 0.4540 & 31.3490 & 0.0062 & 0.9786 \\
                      & 0.0500 & 1.0000 & 1.0000 & 0.6461 & 0.4239 & 34.3950 & 0.0035 & 0.9857 \\
                      & 0.1000 & 1.0000 & 1.0000 & 0.6329 & 0.4093 & 35.3308 & 0.0004 & 1.0000 \\
\midrule
\multirow{3}{*}{Balanced} & 0.0200 & 0.9786 & 0.9857 & 0.5428 & 0.4211 & 22.4251 & 0.0082 & 0.9286 \\
                          & 0.0500 & 0.9929 & 1.0000 & 0.5292 & 0.3972 & 24.9428 & 0.0043 & 0.9786 \\
                          & 0.1000 & 1.0000 & 1.0000 & 0.5123 & 0.3862 & 24.6216 & 0.0016 & 0.9893 \\
\midrule
\multirow{3}{*}{Strong} & 0.0200 & 0.9500 & 0.9286 & 0.4765 & 0.4011 & 15.8365 & 0.0095 & 0.8429 \\
                        & 0.0500 & 0.9893 & 0.9964 & 0.4615 & 0.3824 & 17.1378 & 0.0064 & 0.9214 \\
                        & 0.1000 & 0.9893 & 1.0000 & 0.4499 & 0.3694 & 17.9052 & 0.0027 & 0.9321 \\
\bottomrule
\end{tabular}
\end{table}

\section{Discussion}
\label{sec:discussion}

\subsection{Reframing Underfitting as a Structural Spectral Mismatch}
\label{subsec:discussion_reframing}

Underfitting is often described as a simple shortage of model capacity, but the present results suggest a more structural interpretation in the small-data, representation-constrained regime studied here. In our setting, the key issue is not merely that the model class is restricted, but that the restricted oracle may place a nontrivial portion of its signal in eigendirections that are weakly supported by the representation covariance. Once this happens, standard shrinkage does not act as a neutral complexity control mechanism. It suppresses most strongly the very directions that remain necessary for accurate prediction within the restricted space, thereby creating a mismatch between the geometry of the signal and the geometry of the regularizer. From this perspective, weak-spectrum underfitting should be understood as a form of spectral misalignment: the learner is not simply too small, but is biased against the coordinates that matter most under the available representation bottleneck.

This reframing also clarifies why the empirical results are more informative than a generic observation that negative regularization can sometimes improve test error. The feasibility, bias decomposition, sign-switch, and automatic-selection results together indicate that the relevant question is not whether one should always reduce or reverse regularization, but whether the current shrinkage pattern is structurally misallocated. In particular, the experiments show that substantial degradation can arise even when the total oracle mass in weak directions is not dominant in an absolute sense, because regularization bias concentrates disproportionately in those directions. This means that underfitting can emerge before it appears as an obvious global lack of flexibility, and that diagnosing it requires attention to where predictive signal lies in the spectrum, rather than only to overall model size, sample size, or aggregate complexity measures.

\subsection{Negative Regularization as Controlled Anti-Shrinkage}
\label{subsec:discussion_antishrinkage}

The present framework suggests that negative regularization is better understood as controlled anti-shrinkage than as unrestricted complexity expansion. In the feasible negative interval, moving the regularization parameter leftward does increase effective complexity, but it does so through a precise spectral mechanism rather than through indiscriminate destabilization. The theoretical results show that the adjustment amplifies recovery most strongly in weak eigendirections, namely those that standard positive shrinkage suppresses most aggressively. This is why the role of negative regularization in our setting is more naturally interpreted as a correction of excessive shrinkage bias than as a departure from regularization itself. The point is not to abandon control, but to relax an already conservative estimator in a direction that is aligned with the geometry of the underfitting problem.

This interpretation is also important for understanding what the empirical results do and do not imply. The experiments do not support the naive claim that more negative values are always better, nor do they suggest that performance improves by approaching the spectral boundary as closely as possible. On the contrary, the observed deterioration near the feasibility boundary reinforces the theoretical distinction between meaningful anti-shrinkage and loss of numerical stability. What matters is the existence of an interior region in which shrinkage can be partially reversed while the estimator remains well posed. Within that region, negative regularization functions as a disciplined rebalancing device: it restores directions that were overly attenuated by conservative shrinkage, yet still operates under explicit spectral constraints rather than outside them.

\subsection{Interpreting the Sign-Switch Regime}
\label{subsec:discussion_sign_switch}

A central implication of the sign-switch analysis is that a negative adjustment should not be read as evidence that negative regularization is intrinsically preferable to zero or positive regularization in a universal sense. In the present theory, the sign-switch arises only relative to an already conservative baseline, which means that the relevant question is not whether regularization should exist, but whether the current level of shrinkage has become excessively restrictive for the structure of the problem. This distinction matters because it places the negative region in a corrective, rather than oppositional, role. A negative adjustment does not overturn the logic of regularization; instead, it indicates that the prevailing bias--variance balance has become skewed by too much conservativeness, so that moving leftward can reduce prediction risk by undoing part of that excess shrinkage.

This viewpoint also clarifies why the sign-switch depends jointly on weak-spectrum alignment, baseline shrinkage, and noise level. When the restricted oracle is sufficiently aligned with weak directions, the bias induced by conservative shrinkage can become large enough that a partial reversal is beneficial. When noise is too strong, however, the variance inflation from anti-shrinkage can dominate, and the preferred adjustment remains nonnegative. The experiments reflect exactly this conditional structure: negative adjustment appears most clearly when shrinkage-sensitive signal is present and noise is not overwhelming, and it weakens or disappears when those conditions fail. The sign-switch should therefore be interpreted less as a standalone phenomenon and more as a diagnostic marker of a particular regime, namely one in which underfitting is driven more by over-conservative bias than by variance control alone.

\subsection{What the Experiments Reveal Beyond Performance}
\label{subsec:discussion_beyond_performance}

The empirical results are most informative when read not as isolated performance comparisons, but as a sequence of checks on the underlying mechanism proposed by the theory. The first set of experiments establishes that the negative region is not merely a formal possibility: a nontrivial feasible interval can exist in practice, and moving within that interval changes effective complexity and coefficient magnitude in the direction predicted by the spectral analysis. The second set of experiments then shows that weak-spectrum underfitting is not an abstract construction detached from observable behavior. By varying oracle alignment while holding the covariance geometry fixed, the results make clear that regularization-induced bias is concentrated according to spectral structure rather than according to coefficient mass alone. Taken together, these experiments do more than show that some negative values can perform well; they verify that the proposed interpretation of anti-shrinkage has a concrete and traceable empirical signature.

The later experiments deepen this point by showing that the advantage of negative adjustment is regime dependent and can be recovered by data-driven selection when the underlying structure is strong enough. The sign-switch results confirm that negative adjustment is favored neither uniformly nor mysteriously, but under the specific interaction of conservative baseline shrinkage, weak-spectrum alignment, and noise level predicted by the oracle analysis. The automatic-selection experiments then indicate that this is not only an oracle-level effect: when the negative regime is pronounced, a selector operating over the full negative-capable family can recover much of the same advantage, whereas in weaker or more marginal settings its behavior becomes correspondingly more conservative. The real-data semi-synthetic study is especially important in this respect, because it shows that the phenomenon persists on top of empirical covariance geometries induced by realistic representation bottlenecks, rather than only in idealized toy constructions.

\subsection{Implications for Small-Data Learning Under Representation Bottlenecks}
\label{subsec:discussion_implications}

One broader implication of the present results is that small-data learning should not be viewed exclusively through the lens of variance control. In many practical settings, especially when prediction is performed through a fixed low-dimensional or otherwise constrained representation, the dominant difficulty may arise not from excessive flexibility but from excessive conservativeness. If the available representation already imposes a bottleneck, then additional shrinkage can interact with that bottleneck in a highly nonuniform way, suppressing precisely those directions that remain most important for prediction within the restricted space. This suggests that the standard intuition of small-data learning---namely, that one should respond primarily by adding more regularization---can be incomplete. The more relevant question may be whether the current estimator is already too biased relative to the geometry of the retained signal.

This perspective also has implications for how representation design and regularization are conceptually linked. Under a representation bottleneck, the effect of regularization cannot be understood independently of where the compressed representation places predictive information in the spectrum. A representation that appears stable in a global sense may still induce a setting in which conservative shrinkage removes disproportionately useful structure, particularly when the retained signal is concentrated in weak directions. In that sense, underfitting in small-data regression cannot always be diagnosed from sample size, test error, or nominal model simplicity alone. It depends on the interaction between data scarcity, representational restriction, and the directional bias introduced by shrinkage, which means that controlling learning in such regimes requires attention not only to the amount of regularization, but also to whether regularization is acting against the structure that the representation has made necessary.

\subsection{Limitations and Future Work}
\label{subsec:discussion_limitations_future}

Several limitations of the present formulation should be kept in mind. First, the analysis is developed for regression under a restricted linear representation with a quadratic penalty, which allows the feasible negative interval, spectral anti-shrinkage mechanism, and bias--variance tradeoff to be characterized in a particularly transparent form. Although this setting is appropriate for isolating the central phenomenon of interest, it does not by itself establish that the same conclusions transfer unchanged to richer nonlinear predictors, alternative loss functions, or more general regularizers. Second, the sign-switch analysis is carried out through a baseline-adjusted oracle surrogate that is deliberately stylized in order to separate shrinkage bias reduction from variance inflation. This makes the mechanism analytically visible, but it also means that the theorem should be interpreted as a structural explanation of when negative adjustment can become justified, rather than as a complete description of every finite-sample training procedure. Third, the empirical study, while designed to test the theory from several complementary angles, still relies on synthetic and semi-synthetic constructions in which the signal geometry is controlled more carefully than it would be in many naturally occurring learning problems.

These limitations point directly to several natural directions for future work. One important extension is to move beyond fixed linear bottlenecks and study whether analogous sign-switch behavior can be characterized for richer function classes, generalized linear models, or settings in which the representation itself is learned jointly with the predictor. Another is to investigate broader families of penalties and selection rules, including adaptive or data-dependent forms of anti-shrinkage that may better reflect heterogeneous spectral structure across problems. On the empirical side, it would be valuable to test whether the present mechanism remains detectable in less stylized real-data settings where representation bottlenecks arise from practical preprocessing, architectural compression, or upstream feature learning rather than from controlled construction. A further challenge is to develop diagnostic criteria that can identify bias-dominated conservative regimes directly from data, so that the relevance of negative-capable regularization can be assessed without relying on oracle quantities that are only available in theoretical analysis.

\section{Conclusion}
\label{sec:conclusion}

We studied negative-capable regularization for small-data regression under representation bottlenecks and showed that underfitting in this regime can arise from a structural mismatch between shrinkage and the spectral location of predictive signal. Within a feasible negative interval, negative regularization acts as controlled anti-shrinkage, increasing effective complexity in a structured way and reducing excessive bias when the restricted oracle is aligned with weak directions. Our theory characterized this mechanism through the feasible negative region, weak-spectrum underfitting, and a sign-switch result under conservative baseline shrinkage, while our experiments showed that the predicted benefit of negative adjustment appears in the relevant regimes and can be recovered by automatic selection. These results suggest that, in small-data learning under representation constraints, the key question is not always how much regularization to add, but whether existing conservativeness is already part of the underfitting problem.

\section*{Reproducibility}

To support reproducibility, we will release the full implementation of our method, including the code for data generation and preprocessing, model training, hyperparameter selection, evaluation, and the scripts used to produce the reported results, at \url{https://github.com/AndrewKim1997/negative-regularization}. The repository will also include documentation describing the experimental setup and instructions for reproducing the main results in the paper. For venues that require anonymization or do not permit external links during review, this section and the public repository link can be removed accordingly.

\bibliographystyle{unsrt}  
\bibliography{references}  

\appendix

\section{Additional Notation and Technical Setup}
\label{app:notation_setup}

In this appendix, we collect the notation and technical conventions used throughout the proofs. We continue to work in the restricted linear representation setting introduced in Section~3. Let
\[
\phi : \mathcal{X} \to \mathbb{R}^m
\]
be the fixed representation map, and write
\[
f_\beta(x) = \phi(x)^\top \beta, \qquad \beta \in \mathbb{R}^m .
\]
The population covariance and cross-moment are
\[
\Sigma := \mathbb{E}[\phi(X)\phi(X)^\top], 
\qquad
g := \mathbb{E}[\phi(X)Y],
\]
and their empirical counterparts are
\[
\hat{\Sigma}_n := \frac{1}{n}\Phi^\top \Phi,
\qquad
\hat{g}_n := \frac{1}{n}\Phi^\top Y,
\]
where
\[
\Phi =
\begin{bmatrix}
\phi(x_1)^\top\\
\vdots\\
\phi(x_n)^\top
\end{bmatrix}
\in \mathbb{R}^{n \times m},
\qquad
Y =
\begin{bmatrix}
y_1\\
\vdots\\
y_n
\end{bmatrix}
\in \mathbb{R}^n .
\]

The population-optimal coefficient within the restricted representation family is denoted by
\[
\beta^\dagger \in \arg\min_{\beta \in \mathbb{R}^m}
\mathbb{E}\bigl[(Y - \phi(X)^\top \beta)^2\bigr].
\]
Whenever $\Sigma$ is invertible, this target is unique and satisfies $\beta^\dagger = \Sigma^{-1}g$. The corresponding excess prediction risk is
\[
\mathcal{E}(\beta) := R(\beta) - R(\beta^\dagger),
\qquad
R(\beta) := \mathbb{E}\bigl[(Y - \phi(X)^\top \beta)^2\bigr].
\]
For $\lambda \in \mathbb{R}$ such that the relevant inverse exists, the sample estimator and the population regularized target are written as
\[
\hat{\beta}_\lambda := (\hat{\Sigma}_n + \lambda I_m)^{-1}\hat{g}_n,
\qquad
\beta^{\mathrm{pop}}_\lambda := (\Sigma + \lambda I_m)^{-1}g.
\]

Throughout the theory, we use the eigendecomposition
\[
\Sigma = U \,\mathrm{diag}(\mu_1,\dots,\mu_m) U^\top,
\qquad
\mu_1 \ge \cdots \ge \mu_m > 0,
\]
with orthonormal eigenvectors $U = (u_1,\dots,u_m)$. We expand the restricted oracle as
\[
\beta^\dagger = \sum_{j=1}^m \alpha_j u_j,
\qquad
\alpha_j := u_j^\top \beta^\dagger .
\]
Similarly, when needed, we write
\[
\hat{\Sigma}_n = \hat{U}\,\mathrm{diag}(\hat{\mu}_1,\dots,\hat{\mu}_m)\hat{U}^\top,
\qquad
\hat{\mu}_1 \ge \cdots \ge \hat{\mu}_m \ge 0 .
\]
For a threshold $\kappa \in [\mu_m,\mu_1]$, the weak and strong index sets are
\[
W_\kappa := \{j : \mu_j \le \kappa\},
\qquad
S_\kappa := \{j : \mu_j > \kappa\},
\]
and the corresponding weak and strong components of $\beta^\dagger$ are
\[
\beta^\dagger_{W,\kappa} := \sum_{j \in W_\kappa} \alpha_j u_j,
\qquad
\beta^\dagger_{S,\kappa} := \sum_{j \in S_\kappa} \alpha_j u_j.
\]

We also use the weak-spectrum summary quantities
\[
A_W(\kappa) := \|\beta^\dagger_{W,\kappa}\|_2^2
= \sum_{j \in W_\kappa} \alpha_j^2,
\qquad
A_S(\kappa) := \|\beta^\dagger_{S,\kappa}\|_2^2
= \sum_{j \in S_\kappa} \alpha_j^2,
\]
and
\[
M_W(\kappa) := \|\Sigma^{1/2}\beta^\dagger_{W,\kappa}\|_2^2
= \sum_{j \in W_\kappa} \mu_j \alpha_j^2,
\qquad
M_S(\kappa) := \|\Sigma^{1/2}\beta^\dagger_{S,\kappa}\|_2^2
= \sum_{j \in S_\kappa} \mu_j \alpha_j^2.
\]
Whenever $\beta^\dagger \neq 0$, the weak-spectrum alignment ratio is
\[
\rho_W(\kappa)
:=
\frac{A_W(\kappa)}{A_W(\kappa)+A_S(\kappa)}.
\]

For the sign-switch analysis, we fix a baseline conservativeness level $\tau_0 > 0$ and use the baseline-adjusted parameterization
\[
\eta(\lambda) := \tau_0 + \lambda.
\]
The corresponding oracle surrogate is
\[
\tilde{\beta}_\lambda
:=
(\Sigma + \eta(\lambda) I_m)^{-1}(\Sigma \beta^\dagger + \xi_n),
\]
where $\xi_n$ is a mean-zero noise term with covariance
\[
\mathrm{Cov}(\xi_n) = \frac{\sigma^2}{n}\Sigma.
\]
Its expected excess prediction risk is denoted by
\[
R_n(\lambda;\tau_0)
:=
\mathbb{E}\bigl[\mathcal{E}(\tilde{\beta}_\lambda)\bigr].
\]

For the automatic-selection analysis, we consider a finite candidate set
\[
\Gamma_n \subset [-\tau_0 + \tau,\, L],
\]
where $\tau \in (0,\tau_0)$ and $L>0$ are fixed constants, so that $\eta(\lambda) \ge \tau > 0$ for all $\lambda \in \Gamma_n$. The baseline-adjusted empirical estimator is
\[
\hat{\beta}^{(\tau_0)}_\lambda
:=
(\hat{\Sigma}_n + \eta(\lambda) I_m)^{-1}\hat{g}_n,
\]
with fitted mean vector
\[
\hat{m}_\lambda := \Phi \hat{\beta}^{(\tau_0)}_\lambda = H_\lambda Y,
\]
where
\[
H_\lambda
:=
\Phi(\Phi^\top\Phi + n\eta(\lambda)I_m)^{-1}\Phi^\top
=
\frac{1}{n}\Phi(\hat{\Sigma}_n + \eta(\lambda)I_m)^{-1}\Phi^\top.
\]
Conditional on the design $\Phi$, we write
\[
m_n :=
\begin{bmatrix}
f^\star(x_1)\\
\vdots\\
f^\star(x_n)
\end{bmatrix},
\qquad
Y = m_n + \varepsilon,
\]
with
\[
\mathbb{E}[\varepsilon \mid \Phi] = 0,
\qquad
\mathbb{E}[\varepsilon\varepsilon^\top \mid \Phi] = \sigma^2 I_n.
\]
The conditional prediction risk and its unbiased criterion are
\[
P_n(\lambda)
:=
\frac{1}{n}\mathbb{E}\bigl[\|m_n - \hat{m}_\lambda\|_2^2 \mid \Phi\bigr],
\]
and
\[
\mathrm{Crit}_n(\lambda)
:=
\frac{1}{n}\|Y-\hat{m}_\lambda\|_2^2
+
\frac{2\sigma^2}{n}\mathrm{tr}(H_\lambda)
-
\sigma^2 .
\]
The selected adjustment is then defined by
\[
\hat{\lambda}_n \in \arg\min_{\lambda \in \Gamma_n} \mathrm{Crit}_n(\lambda).
\]

These conventions will be used throughout the appendix without repeated redefinition. In particular, all matrix norms are Euclidean/operator norms according to context, all traces are taken over square matrices of compatible dimension, and all expectations are with respect to the randomness explicitly indicated in the corresponding statement.

\section{Feasible Negative Regularization and Spectral Complexity Expansion}
\label{app:proofs_feasible_complexity}

This subsection collects the proofs underlying the feasible negative interval and the spectral complexity expansion results stated in Section~3.2--Section~3.3. We begin from the quadratic representation of excess risk, then characterize sample and population well-posedness, and finally derive the spectral anti-shrinkage identities that explain why moving into the negative region increases effective complexity most strongly along weak eigendirections.

\subsection{Quadratic excess-risk representation and well-posedness}
\label{app:proofs_wellposedness}

We first verify the quadratic form of the excess risk and the basic spectral characterization of the feasible negative interval.

For any $\beta \in \mathbb{R}^m$, write $\phi=\phi(X)$ for brevity. Expanding around the restricted oracle $\beta^\dagger$ gives
\[
Y-\phi^\top \beta
=
\bigl(Y-\phi^\top \beta^\dagger\bigr)-\phi^\top(\beta-\beta^\dagger).
\]
Therefore,
\begin{align*}
R(\beta)
&=
\mathbb{E}\bigl[(Y-\phi^\top\beta)^2\bigr] \\
&=
\mathbb{E}\bigl[(Y-\phi^\top\beta^\dagger)^2\bigr]
-2\,\mathbb{E}\bigl[(Y-\phi^\top\beta^\dagger)\phi^\top(\beta-\beta^\dagger)\bigr]
+\mathbb{E}\bigl[(\phi^\top(\beta-\beta^\dagger))^2\bigr].
\end{align*}
Since $\beta^\dagger$ satisfies the population normal equation $\Sigma \beta^\dagger=g$, the cross term vanishes:
\[
\mathbb{E}\bigl[(Y-\phi^\top\beta^\dagger)\phi\bigr]
=
\mathbb{E}[\phi Y]-\mathbb{E}[\phi\phi^\top]\beta^\dagger
=
g-\Sigma\beta^\dagger
=
0.
\]
Hence
\[
R(\beta)
=
R(\beta^\dagger)+(\beta-\beta^\dagger)^\top \Sigma (\beta-\beta^\dagger),
\]
and so
\[
\mathcal{E}(\beta)
=
R(\beta)-R(\beta^\dagger)
=
(\beta-\beta^\dagger)^\top \Sigma (\beta-\beta^\dagger).
\]
This proves the quadratic excess-risk representation. In particular, if $\Sigma \succ 0$, then $\beta^\dagger$ is the unique minimizer of $R(\beta)$ over $\mathbb{R}^m$.

We now turn to sample well-posedness. Recall
\[
Q_n(\beta;\lambda)
=
\frac{1}{n}\|Y-\Phi\beta\|_2^2+\lambda\|\beta\|_2^2
=
\frac{1}{n}\|Y\|_2^2-2\hat g_n^\top \beta+\beta^\top(\hat\Sigma_n+\lambda I_m)\beta.
\]
Its Hessian is
\[
\nabla_\beta^2 Q_n(\beta;\lambda)=2(\hat\Sigma_n+\lambda I_m).
\]
Hence $Q_n(\cdot;\lambda)$ is strictly convex if and only if $\hat\Sigma_n+\lambda I_m\succ 0$. In that case the objective is coercive and admits a unique minimizer. Conversely, if $\hat\Sigma_n+\lambda I_m$ is not positive definite, then there exists $v\neq 0$ such that
\[
v^\top(\hat\Sigma_n+\lambda I_m)v\le 0.
\]
If the inequality is strict, then
\[
Q_n(tv;\lambda)\to -\infty
\qquad \text{as } |t|\to\infty,
\]
so the objective is unbounded below. If equality holds, strict convexity fails, and uniqueness is lost in general. Therefore the following are equivalent:
\[
Q_n(\cdot;\lambda)\text{ is bounded below and has a unique minimizer}
\iff
\hat\Sigma_n+\lambda I_m\succ 0.
\]

Let $\hat\mu_1\ge \cdots \ge \hat\mu_m\ge 0$ be the eigenvalues of $\hat\Sigma_n$. The eigenvalues of $\hat\Sigma_n+\lambda I_m$ are exactly $\hat\mu_j+\lambda$, $j=1,\dots,m$. Thus
\[
\hat\Sigma_n+\lambda I_m\succ 0
\iff
\hat\mu_j+\lambda>0\ \text{for all }j
\iff
\lambda>-\hat\mu_m.
\]
Whenever this holds, the first-order condition
\[
-2\hat g_n+2(\hat\Sigma_n+\lambda I_m)\beta=0
\]
has the unique solution
\[
\hat\beta_\lambda=(\hat\Sigma_n+\lambda I_m)^{-1}\hat g_n.
\]
Moreover, since the smallest eigenvalue of $\hat\Sigma_n+\lambda I_m$ is $\hat\mu_m+\lambda$,
\[
\|(\hat\Sigma_n+\lambda I_m)^{-1}\|_{\mathrm{op}}
=
\frac{1}{\hat\mu_m+\lambda},
\qquad
\|\hat\beta_\lambda\|_2
\le
\frac{\|\hat g_n\|_2}{\hat\mu_m+\lambda}.
\]

The same argument yields the population counterpart. Since
\[
Q^{\mathrm{pop}}(\beta;\lambda)
=
R(\beta)+\lambda\|\beta\|_2^2
=
R(\beta^\dagger)+(\beta-\beta^\dagger)^\top\Sigma(\beta-\beta^\dagger)+\lambda\|\beta\|_2^2,
\]
differentiation gives
\[
\nabla_\beta Q^{\mathrm{pop}}(\beta;\lambda)
=
2(\Sigma+\lambda I_m)\beta-2\Sigma\beta^\dagger
=
2(\Sigma+\lambda I_m)\beta-2g.
\]
Thus, whenever $\Sigma+\lambda I_m\succ 0$,
\[
\beta_\lambda^{\mathrm{pop}}
=
(\Sigma+\lambda I_m)^{-1}g
=
(\Sigma+\lambda I_m)^{-1}\Sigma\beta^\dagger.
\]
If $\mu_1\ge \cdots \ge \mu_m>0$ are the eigenvalues of $\Sigma$, then
\[
\Sigma+\lambda I_m\succ 0
\iff
\lambda>-\mu_m.
\]
Hence the population feasible negative interval is $( -\mu_m,0 )$, and
\[
\|(\Sigma+\lambda I_m)^{-1}\|_{\mathrm{op}}
=
\frac{1}{\mu_m+\lambda},
\qquad
\|\beta_\lambda^{\mathrm{pop}}\|_2
\le
\frac{\|g\|_2}{\mu_m+\lambda}.
\]

Finally, if $\tau\in(0,\hat\mu_m)$ and $\lambda\in[-\hat\mu_m+\tau,0)$, then $\hat\mu_m+\lambda\ge \tau$, so
\[
\|(\hat\Sigma_n+\lambda I_m)^{-1}\|_{\mathrm{op}}\le \frac{1}{\tau},
\qquad
\|\hat\beta_\lambda\|_2\le \frac{\|\hat g_n\|_2}{\tau}.
\]
Moreover, by the resolvent identity,
\[
(\hat\Sigma_n+\lambda_1 I_m)^{-1}-(\hat\Sigma_n+\lambda_2 I_m)^{-1}
=
(\lambda_2-\lambda_1)
(\hat\Sigma_n+\lambda_1 I_m)^{-1}
(\hat\Sigma_n+\lambda_2 I_m)^{-1},
\]
and therefore
\[
\hat\beta_{\lambda_1}-\hat\beta_{\lambda_2}
=
(\lambda_2-\lambda_1)
(\hat\Sigma_n+\lambda_1 I_m)^{-1}
(\hat\Sigma_n+\lambda_2 I_m)^{-1}\hat g_n.
\]
Taking norms yields
\[
\|\hat\beta_{\lambda_1}-\hat\beta_{\lambda_2}\|_2
\le
\frac{|\lambda_1-\lambda_2|}{\tau^2}\|\hat g_n\|_2.
\]

\subsection{Population regularized target and spectral filter form}
\label{app:proofs_population_filter}

We next derive the spectral representation of the population regularized target. Let
\[
\Sigma
=
U\,\mathrm{diag}(\mu_1,\dots,\mu_m)U^\top,
\qquad
U=(u_1,\dots,u_m),
\]
with $\mu_1\ge \cdots \ge \mu_m>0$, and expand
\[
\beta^\dagger=\sum_{j=1}^m \alpha_j u_j,
\qquad
\alpha_j=u_j^\top \beta^\dagger.
\]
Then
\[
\Sigma+\lambda I_m
=
U\,\mathrm{diag}(\mu_1+\lambda,\dots,\mu_m+\lambda)U^\top,
\]
and hence
\[
(\Sigma+\lambda I_m)^{-1}
=
U\,\mathrm{diag}\!\left(\frac{1}{\mu_1+\lambda},\dots,\frac{1}{\mu_m+\lambda}\right)U^\top.
\]
Multiplying by $\Sigma$ gives
\[
(\Sigma+\lambda I_m)^{-1}\Sigma
=
U\,\mathrm{diag}\!\left(\frac{\mu_1}{\mu_1+\lambda},\dots,\frac{\mu_m}{\mu_m+\lambda}\right)U^\top.
\]
Therefore
\[
\beta_\lambda^{\mathrm{pop}}
=
(\Sigma+\lambda I_m)^{-1}\Sigma\beta^\dagger
=
\sum_{j=1}^m \frac{\mu_j}{\mu_j+\lambda}\alpha_j u_j.
\]
Subtracting $\beta^\dagger=\sum_{j=1}^m \alpha_j u_j$, we obtain
\[
\beta_\lambda^{\mathrm{pop}}-\beta^\dagger
=
\sum_{j=1}^m
\left(
\frac{\mu_j}{\mu_j+\lambda}-1
\right)\alpha_j u_j
=
-\sum_{j=1}^m \frac{\lambda}{\mu_j+\lambda}\alpha_j u_j.
\]
Equivalently,
\[
\beta_\lambda^{\mathrm{pop}}-\beta^\dagger
=
-\lambda(\Sigma+\lambda I_m)^{-1}\beta^\dagger.
\]

Thus each oracle coordinate $\alpha_j$ is rescaled by the spectral factor
\[
s_j(\lambda):=\frac{\mu_j}{\mu_j+\lambda}.
\]
This is the basic spectral filter associated with negative-capable ridge regularization.

\subsection{Anti-shrinkage monotonicity and effective complexity increase}
\label{app:proofs_antishrinkage_complexity}

For each fixed $j$, define
\[
s_j(\lambda)=\frac{\mu_j}{\mu_j+\lambda},
\qquad
\lambda>-\mu_j.
\]
Direct differentiation gives
\[
\frac{d}{d\lambda}s_j(\lambda)
=
-\frac{\mu_j}{(\mu_j+\lambda)^2}
<
0.
\]
Hence $s_j(\lambda)$ is strictly decreasing in $\lambda$. In particular,
\[
s_j(0)=1,
\]
so if $\lambda>0$, then $0<s_j(\lambda)<1$, while if $-\mu_j<\lambda<0$, then $s_j(\lambda)>1$. Therefore moving $\lambda$ into the negative region expands each spectral coordinate away from zero.

To compare different eigendirections, fix $\lambda<0$ and define
\[
h_\lambda(\mu):=\frac{\mu}{\mu+\lambda}.
\]
Then
\[
h_\lambda'(\mu)
=
\frac{\lambda}{(\mu+\lambda)^2}
<
0.
\]
Thus $h_\lambda(\mu)$ is strictly decreasing in $\mu$. Consequently, if $\mu_i\ge \mu_j$, then
\[
s_i(\lambda)\le s_j(\lambda),
\]
with strict inequality whenever $\mu_i>\mu_j$. This proves that negative regularization amplifies weaker eigendirections more strongly.

We now compute the resulting effective complexity. Define the population effective degrees of freedom by
\[
df(\lambda):=\mathrm{tr}\bigl(\Sigma(\Sigma+\lambda I_m)^{-1}\bigr),
\qquad
\lambda>-\mu_m.
\]
Using the eigendecomposition,
\[
\Sigma(\Sigma+\lambda I_m)^{-1}
=
U\,\mathrm{diag}\!\left(
\frac{\mu_1}{\mu_1+\lambda},\dots,\frac{\mu_m}{\mu_m+\lambda}
\right)U^\top,
\]
and therefore
\[
df(\lambda)
=
\sum_{j=1}^m \frac{\mu_j}{\mu_j+\lambda}.
\]
Differentiating termwise yields
\[
\frac{d}{d\lambda}df(\lambda)
=
-\sum_{j=1}^m \frac{\mu_j}{(\mu_j+\lambda)^2}
<
0.
\]
Hence $df(\lambda)$ is strictly decreasing in $\lambda$. It follows immediately that
\[
\lambda>0 \ \Rightarrow\  df(\lambda)<m,
\qquad
\lambda=0 \ \Rightarrow\  df(0)=m,
\qquad
-\mu_m<\lambda<0 \ \Rightarrow\  df(\lambda)>m.
\]
Moreover, as $\lambda\downarrow -\mu_m$, the term $\mu_m/(\mu_m+\lambda)$ diverges, so
\[
df(\lambda)\to\infty.
\]

The empirical counterpart is identical. If
\[
\hat\Sigma_n
=
\hat U\,\mathrm{diag}(\hat\mu_1,\dots,\hat\mu_m)\hat U^\top,
\]
define
\[
\widehat{df}_n(\lambda)
:=
\mathrm{tr}\bigl(\hat\Sigma_n(\hat\Sigma_n+\lambda I_m)^{-1}\bigr),
\qquad
\lambda>-\hat\mu_m.
\]
Then
\[
\widehat{df}_n(\lambda)
=
\sum_{j=1}^m \frac{\hat\mu_j}{\hat\mu_j+\lambda},
\qquad
\frac{d}{d\lambda}\widehat{df}_n(\lambda)
=
-\sum_{j=1}^m \frac{\hat\mu_j}{(\hat\mu_j+\lambda)^2}
<
0.
\]
In particular, for any $\lambda_1<\lambda_2$ with both values in the empirical well-posed region,
\[
\widehat{df}_n(\lambda_1)>\widehat{df}_n(\lambda_2).
\]

\subsection{Distance from the restricted oracle and related identities}
\label{app:proofs_oracle_distance}

Finally, we record a useful identity for the deviation of the population regularized target from the restricted oracle. Starting from
\[
\beta_\lambda^{\mathrm{pop}}
=
(\Sigma+\lambda I_m)^{-1}\Sigma\beta^\dagger,
\]
use the decomposition
\[
\Sigma=(\Sigma+\lambda I_m)-\lambda I_m.
\]
Then
\begin{align*}
\beta_\lambda^{\mathrm{pop}}
&=
(\Sigma+\lambda I_m)^{-1}\bigl((\Sigma+\lambda I_m)-\lambda I_m\bigr)\beta^\dagger \\
&=
\beta^\dagger-\lambda(\Sigma+\lambda I_m)^{-1}\beta^\dagger,
\end{align*}
and therefore
\[
\beta_\lambda^{\mathrm{pop}}-\beta^\dagger
=
-\lambda(\Sigma+\lambda I_m)^{-1}\beta^\dagger.
\]
Expanding in the eigenbasis,
\[
(\Sigma+\lambda I_m)^{-1}\beta^\dagger
=
\sum_{j=1}^m \frac{\alpha_j}{\mu_j+\lambda}u_j,
\]
so
\[
\beta_\lambda^{\mathrm{pop}}-\beta^\dagger
=
-\sum_{j=1}^m \frac{\lambda}{\mu_j+\lambda}\alpha_j u_j.
\]
Taking squared Euclidean norms and using orthonormality of $\{u_j\}_{j=1}^m$ gives
\[
\|\beta_\lambda^{\mathrm{pop}}-\beta^\dagger\|_2^2
=
\sum_{j=1}^m \frac{\lambda^2}{(\mu_j+\lambda)^2}\alpha_j^2.
\]

Together with the previous subsection, this identity makes the anti-shrinkage mechanism explicit. Inside the feasible negative interval, decreasing $\lambda$ expands the oracle coordinates through the factors $\mu_j/(\mu_j+\lambda)$, with stronger amplification on smaller eigenvalues, and therefore increases effective complexity in a directionally structured rather than uniform manner.

\section{Weak-Spectrum Underfitting and Sign-Switch Behavior}
\label{app:proofs_weak_spectrum_signswitch}

This subsection collects the proofs for the weak-spectrum decomposition, the induced bias concentration in weak eigendirections, and the sign-switch result under conservative baseline shrinkage. The central point is that once the restricted oracle places enough mass on weak directions, shrinkage bias becomes spectrally concentrated there, which in turn creates a regime where a negative adjustment can improve the oracle criterion relative to a conservative positive baseline.

\subsection{Weak--strong decomposition and spectral sandwich bounds}
\label{app:proofs_weakstrong_decomposition}

Fix a threshold $\kappa \in [\mu_m,\mu_1]$ and recall the weak and strong index sets
\[
W_\kappa := \{j : \mu_j \le \kappa\},
\qquad
S_\kappa := \{j : \mu_j > \kappa\}.
\]
Using the eigendecomposition
\[
\Sigma = U\,\mathrm{diag}(\mu_1,\dots,\mu_m)U^\top,
\qquad
\beta^\dagger = \sum_{j=1}^m \alpha_j u_j,
\]
we write
\[
\beta^\dagger_{W,\kappa} := \sum_{j\in W_\kappa}\alpha_j u_j,
\qquad
\beta^\dagger_{S,\kappa} := \sum_{j\in S_\kappa}\alpha_j u_j.
\]
By orthonormality,
\[
A_W(\kappa)
=
\|\beta^\dagger_{W,\kappa}\|_2^2
=
\sum_{j\in W_\kappa}\alpha_j^2,
\qquad
A_S(\kappa)
=
\|\beta^\dagger_{S,\kappa}\|_2^2
=
\sum_{j\in S_\kappa}\alpha_j^2.
\]
Likewise,
\[
M_W(\kappa)
=
\|\Sigma^{1/2}\beta^\dagger_{W,\kappa}\|_2^2
=
\sum_{j\in W_\kappa}\mu_j\alpha_j^2,
\qquad
M_S(\kappa)
=
\|\Sigma^{1/2}\beta^\dagger_{S,\kappa}\|_2^2
=
\sum_{j\in S_\kappa}\mu_j\alpha_j^2.
\]

Since every $j\in W_\kappa$ satisfies $\mu_m \le \mu_j \le \kappa$, we obtain
\[
\mu_m\sum_{j\in W_\kappa}\alpha_j^2
\le
\sum_{j\in W_\kappa}\mu_j\alpha_j^2
\le
\kappa\sum_{j\in W_\kappa}\alpha_j^2,
\]
that is,
\[
\mu_m A_W(\kappa)\le M_W(\kappa)\le \kappa A_W(\kappa).
\]
Similarly, every $j\in S_\kappa$ satisfies $\kappa < \mu_j \le \mu_1$, so
\[
\kappa\sum_{j\in S_\kappa}\alpha_j^2
<
\sum_{j\in S_\kappa}\mu_j\alpha_j^2
\le
\mu_1\sum_{j\in S_\kappa}\alpha_j^2,
\]
hence
\[
\kappa A_S(\kappa)<M_S(\kappa)\le \mu_1 A_S(\kappa).
\]
Equivalently,
\[
\|\Sigma^{1/2}\beta^\dagger_{W,\kappa}\|_2^2
\le
\kappa\|\beta^\dagger_{W,\kappa}\|_2^2,
\qquad
\|\Sigma^{1/2}\beta^\dagger_{S,\kappa}\|_2^2
>
\kappa\|\beta^\dagger_{S,\kappa}\|_2^2.
\]

These inequalities make explicit the geometric asymmetry underlying weak-spectrum underfitting: Euclidean oracle mass placed in weak directions carries relatively small covariance-weighted mass, which is precisely why shrinkage can distort the coefficient geometry substantially before that distortion is fully reflected in the prediction norm.

\subsection{Bias decomposition and weak-direction sensitivity}
\label{app:proofs_bias_decomposition}

We now compute the regularization-induced population bias and show that, in the negative region, it is spectrally largest on weak directions.

Recall from Appendix~\ref{app:proofs_feasible_complexity} that
\[
\beta_\lambda^{\mathrm{pop}}-\beta^\dagger
=
-\sum_{j=1}^m \frac{\lambda}{\mu_j+\lambda}\alpha_j u_j,
\qquad
\lambda>-\mu_m.
\]
By the quadratic excess-risk representation,
\[
B(\lambda)
:=
\mathcal{E}(\beta_\lambda^{\mathrm{pop}})
=
(\beta_\lambda^{\mathrm{pop}}-\beta^\dagger)^\top
\Sigma
(\beta_\lambda^{\mathrm{pop}}-\beta^\dagger).
\]
Substituting the spectral expansion yields
\[
B(\lambda)
=
\sum_{j=1}^m
\mu_j
\frac{\lambda^2}{(\mu_j+\lambda)^2}
\alpha_j^2.
\]
Splitting this sum over the weak and strong index sets gives
\[
B(\lambda)=B_W(\lambda;\kappa)+B_S(\lambda;\kappa),
\]
where
\[
B_W(\lambda;\kappa)
:=
\sum_{j\in W_\kappa}
\mu_j
\frac{\lambda^2}{(\mu_j+\lambda)^2}
\alpha_j^2,
\qquad
B_S(\lambda;\kappa)
:=
\sum_{j\in S_\kappa}
\mu_j
\frac{\lambda^2}{(\mu_j+\lambda)^2}
\alpha_j^2.
\]

Define the directional bias multiplier
\[
b_\lambda(\mu)
:=
\mu\frac{\lambda^2}{(\mu+\lambda)^2}.
\]
Differentiating with respect to $\mu$ gives
\[
b_\lambda'(\mu)
=
\lambda^2\frac{\lambda-\mu}{(\mu+\lambda)^3}.
\]
If $\lambda<0$ and $\mu>-\lambda$, then $\mu+\lambda>0$ and $\lambda-\mu<0$, so
\[
b_\lambda'(\mu)<0.
\]
Thus for every fixed $\lambda\in(-\mu_m,0)$, the map $\mu\mapsto b_\lambda(\mu)$ is strictly decreasing. Consequently, if $\mu_i\ge \mu_j$, then
\[
b_\lambda(\mu_i)\le b_\lambda(\mu_j),
\]
with strict inequality whenever $\mu_i>\mu_j$.

This yields the ordering used in the main text. Writing
\[
b_j(\lambda):=\mu_j\frac{\lambda^2}{(\mu_j+\lambda)^2},
\]
we have, for $\lambda\in(-\mu_m,0)$,
\[
\mu_i\ge \mu_j
\quad\Longrightarrow\quad
b_i(\lambda)\le b_j(\lambda).
\]
Therefore
\[
B_W(\lambda;\kappa)
=
\sum_{j\in W_\kappa} b_j(\lambda)\alpha_j^2
\ge
\Bigl(\min_{j\in W_\kappa} b_j(\lambda)\Bigr)
\sum_{j\in W_\kappa}\alpha_j^2
=
\Bigl(\min_{j\in W_\kappa} b_j(\lambda)\Bigr)A_W(\kappa),
\]
and similarly
\[
B_S(\lambda;\kappa)
=
\sum_{j\in S_\kappa} b_j(\lambda)\alpha_j^2
\le
\Bigl(\max_{j\in S_\kappa} b_j(\lambda)\Bigr)
A_S(\kappa).
\]
If both index sets are nonempty, then every eigenvalue in $W_\kappa$ is at most $\kappa$, while every eigenvalue in $S_\kappa$ is strictly larger than $\kappa$. Since $b_\lambda(\mu)$ is decreasing in $\mu$ for $\lambda<0$,
\[
\min_{j\in W_\kappa} b_j(\lambda)
\ge
\max_{j\in S_\kappa} b_j(\lambda).
\]

In particular,
\[
B(\lambda)\ge B_W(\lambda;\kappa)
\ge
\Bigl(\min_{j\in W_\kappa} b_j(\lambda)\Bigr)A_W(\kappa).
\]
Under the stronger weak-spectrum alignment condition
\[
A_W(\kappa)\ge c_0\|\beta^\dagger\|_2^2
\qquad\text{for some } c_0\in(0,1],
\]
this implies
\[
B(\lambda)
\ge
c_0
\Bigl(\min_{j\in W_\kappa} b_j(\lambda)\Bigr)
\|\beta^\dagger\|_2^2.
\]

Thus, once a nonnegligible fraction of the restricted oracle lies in weak eigendirections, the negative-capable family inherits a corresponding lower bound on regularization bias from those weak directions. This is the structural mechanism behind weak-spectrum underfitting.

\subsection{Local oracle criterion under conservative baseline shrinkage}
\label{app:proofs_oracle_criterion}

A genuine sign-switch cannot be read off directly from $\beta_\lambda^{\mathrm{pop}}$, since the excess risk $\mathcal{E}(\beta_\lambda^{\mathrm{pop}})$ is minimized at $\lambda=0$. The sign question therefore has to be posed relative to an already conservative baseline.

Fix $\tau_0>0$ and define
\[
\eta(\lambda):=\tau_0+\lambda.
\]
We consider the baseline-adjusted oracle surrogate
\[
\tilde\beta_\lambda
=
(\Sigma+\eta(\lambda)I_m)^{-1}(\Sigma\beta^\dagger+\xi_n),
\]
where
\[
\mathbb{E}[\xi_n]=0,
\qquad
\mathrm{Cov}(\xi_n)=\frac{\sigma^2}{n}\Sigma.
\]
Its expected excess risk is
\[
R_n(\lambda;\tau_0)
:=
\mathbb{E}\bigl[\mathcal{E}(\tilde\beta_\lambda)\bigr].
\]

Expand everything in the eigenbasis of $\Sigma$. Writing
\[
\beta^\dagger=\sum_{j=1}^m \alpha_j u_j,
\qquad
\zeta_j:=u_j^\top\xi_n,
\]
we obtain
\[
\tilde\beta_\lambda
=
(\Sigma+\eta I_m)^{-1}\Sigma\beta^\dagger
+
(\Sigma+\eta I_m)^{-1}\xi_n
=
\sum_{j=1}^m
\left(
\frac{\mu_j}{\mu_j+\eta}\alpha_j
+
\frac{\zeta_j}{\mu_j+\eta}
\right)u_j,
\]
where $\eta=\eta(\lambda)$ for brevity. Therefore
\[
\tilde\beta_\lambda-\beta^\dagger
=
\sum_{j=1}^m
\left(
-\frac{\eta}{\mu_j+\eta}\alpha_j
+
\frac{\zeta_j}{\mu_j+\eta}
\right)u_j.
\]
Using the quadratic excess-risk identity,
\[
\mathcal{E}(\tilde\beta_\lambda)
=
(\tilde\beta_\lambda-\beta^\dagger)^\top
\Sigma
(\tilde\beta_\lambda-\beta^\dagger)
=
\sum_{j=1}^m
\mu_j
\left(
-\frac{\eta}{\mu_j+\eta}\alpha_j
+
\frac{\zeta_j}{\mu_j+\eta}
\right)^2.
\]
Taking expectations and using $\mathbb{E}[\zeta_j]=0$, we get
\[
R_n(\lambda;\tau_0)
=
\sum_{j=1}^m
\mu_j\frac{\eta(\lambda)^2\alpha_j^2}{(\mu_j+\eta(\lambda))^2}
+
\sum_{j=1}^m
\mu_j\frac{\mathbb{E}[\zeta_j^2]}{(\mu_j+\eta(\lambda))^2}.
\]
Since
\[
\mathbb{E}[\zeta_j^2]
=
u_j^\top \mathrm{Cov}(\xi_n)u_j
=
\frac{\sigma^2}{n}u_j^\top\Sigma u_j
=
\frac{\sigma^2}{n}\mu_j,
\]
it follows that
\[
R_n(\lambda;\tau_0)
=
\sum_{j=1}^m
\mu_j\frac{\eta(\lambda)^2\alpha_j^2}{(\mu_j+\eta(\lambda))^2}
+
\frac{\sigma^2}{n}
\sum_{j=1}^m
\frac{\mu_j^2}{(\mu_j+\eta(\lambda))^2}.
\]
Thus
\[
R_n(\lambda;\tau_0)=B_n(\lambda;\tau_0)+V_n(\lambda;\tau_0),
\]
where
\[
B_n(\lambda;\tau_0)
:=
\sum_{j=1}^m
\mu_j\frac{\eta(\lambda)^2\alpha_j^2}{(\mu_j+\eta(\lambda))^2},
\qquad
V_n(\lambda;\tau_0)
:=
\frac{\sigma^2}{n}
\sum_{j=1}^m
\frac{\mu_j^2}{(\mu_j+\eta(\lambda))^2}.
\]

We now differentiate with respect to $\lambda$. Since $\eta'(\lambda)=1$, it is enough to differentiate with respect to $\eta$. For the bias term,
\[
\frac{\partial}{\partial\eta}
\left(
\mu_j\frac{\eta^2\alpha_j^2}{(\mu_j+\eta)^2}
\right)
=
2\mu_j^2\frac{\eta\alpha_j^2}{(\mu_j+\eta)^3}.
\]
For the variance term,
\[
\frac{\partial}{\partial\eta}
\left(
\frac{\sigma^2}{n}\frac{\mu_j^2}{(\mu_j+\eta)^2}
\right)
=
-2\frac{\sigma^2}{n}\frac{\mu_j^2}{(\mu_j+\eta)^3}.
\]
Summing over $j$ yields
\[
\frac{\partial}{\partial\lambda}R_n(\lambda;\tau_0)
=
2\sum_{j=1}^m
\frac{\mu_j^2}{(\mu_j+\eta(\lambda))^3}
\left(
\eta(\lambda)\alpha_j^2-\frac{\sigma^2}{n}
\right).
\]
In particular, at the baseline point $\lambda=0$,
\[
\frac{\partial}{\partial\lambda}R_n(0;\tau_0)
=
2\sum_{j=1}^m
\frac{\mu_j^2}{(\mu_j+\tau_0)^3}
\left(
\tau_0\alpha_j^2-\frac{\sigma^2}{n}
\right).
\]

This expression isolates the sign-switch mechanism precisely: the derivative is positive when the reduction in baseline shrinkage bias dominates the induced variance increase.

\subsection{Sign-switch theorem and weak-spectrum sufficient conditions}
\label{app:proofs_signswitch_main}

We now prove the local sign-switch statement. Suppose
\[
\frac{\partial}{\partial\lambda}R_n(0;\tau_0)>0.
\]
Since $R_n(\lambda;\tau_0)$ is differentiable at $\lambda=0$, there exists $\delta\in(0,\tau_0)$ such that
\[
R_n(-\delta;\tau_0)<R_n(0;\tau_0).
\]
Therefore the minimum of $R_n(\lambda;\tau_0)$ over the feasible adjustment range must be strictly smaller than its value at $\lambda=0$. Consequently, at least one oracle minimizer
\[
\lambda_n^\star(\tau_0)\in
\arg\min_{\lambda>-\tau_0-\mu_m}R_n(\lambda;\tau_0)
\]
satisfies
\[
\lambda_n^\star(\tau_0)<0.
\]
This proves the sign-switch theorem.

To express the derivative condition in weak-spectrum form, define
\[
w_j(\tau_0):=\frac{\mu_j^2}{(\mu_j+\tau_0)^3}.
\]
Then the derivative at the baseline point can be written as
\[
\frac{\partial}{\partial\lambda}R_n(0;\tau_0)
=
2\tau_0\sum_{j=1}^m w_j(\tau_0)\alpha_j^2
-
2\frac{\sigma^2}{n}\sum_{j=1}^m w_j(\tau_0).
\]
Discarding the nonnegative contribution from $S_\kappa$, we obtain
\[
\sum_{j=1}^m w_j(\tau_0)\alpha_j^2
\ge
\sum_{j\in W_\kappa} w_j(\tau_0)\alpha_j^2
\ge
\Bigl(\min_{j\in W_\kappa}w_j(\tau_0)\Bigr)
\sum_{j\in W_\kappa}\alpha_j^2.
\]
Hence
\[
\frac{\partial}{\partial\lambda}R_n(0;\tau_0)
\ge
2\tau_0
\Bigl(\min_{j\in W_\kappa}w_j(\tau_0)\Bigr)
A_W(\kappa)
-
2\frac{\sigma^2}{n}\sum_{j=1}^m w_j(\tau_0).
\]
Therefore, if
\[
\tau_0
\Bigl(\min_{j\in W_\kappa}w_j(\tau_0)\Bigr)
A_W(\kappa)
>
\frac{\sigma^2}{n}\sum_{j=1}^m w_j(\tau_0),
\]
then
\[
\frac{\partial}{\partial\lambda}R_n(0;\tau_0)>0,
\]
and the sign-switch theorem applies.

Under the stronger alignment condition
\[
A_W(\kappa)\ge c_0\|\beta^\dagger\|_2^2
\qquad\text{for some }c_0\in(0,1],
\]
it is enough that
\[
\tau_0 c_0
\Bigl(\min_{j\in W_\kappa}w_j(\tau_0)\Bigr)
\|\beta^\dagger\|_2^2
>
\frac{\sigma^2}{n}\sum_{j=1}^m w_j(\tau_0).
\]
Indeed, this implies
\[
\tau_0
\Bigl(\min_{j\in W_\kappa}w_j(\tau_0)\Bigr)
A_W(\kappa)
>
\frac{\sigma^2}{n}\sum_{j=1}^m w_j(\tau_0),
\]
and hence again
\[
\frac{\partial}{\partial\lambda}R_n(0;\tau_0)>0.
\]
Therefore there exists $\delta\in(0,\tau_0)$ such that
\[
R_n(-\delta;\tau_0)<R_n(0;\tau_0),
\]
and at least one oracle adjustment lies in the negative region.

Taken together, these arguments show that the sign-switch is not an isolated algebraic curiosity. It is a structural consequence of three ingredients acting jointly: a conservative baseline $\tau_0$, sufficient restricted-oracle mass in weak eigendirections, and a noise level small enough that the bias relief from moving leftward outweighs the corresponding variance inflation.

\section{Criterion-Based Automatic Selection over the Negative-Capable Family}
\label{app:proofs_selection}

This subsection collects the proofs for the criterion-based automatic selection results in Section~3.6. We first verify the conditional unbiasedness of the empirical criterion, then derive the oracle inequality for the selected adjustment, and finally show how a strict negative-region margin forces the selector to recover a negative adjustment. We conclude with the approximation argument that links optimization over a finite candidate grid to the corresponding continuous negative-capable interval.

\subsection{Conditional unbiased risk estimation}
\label{app:proofs_unbiased_risk}

Fix $\lambda \in \Gamma_n$ and abbreviate
\[
H := H_\lambda,
\qquad
\hat m_\lambda = H_\lambda Y = HY.
\]
Recall that, conditional on the design $\Phi$,
\[
Y = m_n + \varepsilon,
\qquad
\mathbb{E}[\varepsilon \mid \Phi] = 0,
\qquad
\mathbb{E}[\varepsilon \varepsilon^\top \mid \Phi] = \sigma^2 I_n,
\]
and that the conditional prediction risk is
\[
P_n(\lambda)
=
\frac{1}{n}\mathbb{E}\bigl[\|m_n-\hat m_\lambda\|_2^2 \mid \Phi\bigr].
\]
The empirical criterion is
\[
\mathrm{Crit}_n(\lambda)
=
\frac{1}{n}\|Y-\hat m_\lambda\|_2^2
+
\frac{2\sigma^2}{n}\mathrm{tr}(H_\lambda)
-
\sigma^2.
\]

We first compute the conditional expectation of the residual term. Since
\[
Y-HY = (I_n-H)m_n + (I_n-H)\varepsilon,
\]
we have
\begin{align*}
\mathbb{E}\bigl[\|Y-HY\|_2^2 \mid \Phi\bigr]
&=
\|(I_n-H)m_n\|_2^2
+
\mathbb{E}\bigl[\varepsilon^\top (I_n-H)^\top (I_n-H)\varepsilon \mid \Phi\bigr].
\end{align*}
Because $\eta(\lambda)>0$, the matrix
\[
H_\lambda
=
\Phi(\Phi^\top\Phi+n\eta(\lambda)I_m)^{-1}\Phi^\top
\]
is symmetric, so
\[
(I_n-H)^\top (I_n-H) = (I_n-H)^2.
\]
Using $\mathbb{E}[\varepsilon\varepsilon^\top \mid \Phi]=\sigma^2 I_n$, we obtain
\[
\mathbb{E}\bigl[\varepsilon^\top (I_n-H)^2\varepsilon \mid \Phi\bigr]
=
\sigma^2 \mathrm{tr}\bigl((I_n-H)^2\bigr).
\]
Hence
\[
\mathbb{E}\bigl[\|Y-HY\|_2^2 \mid \Phi\bigr]
=
\|(I_n-H)m_n\|_2^2
+
\sigma^2 \mathrm{tr}\bigl((I_n-H)^2\bigr).
\]

Now expand
\[
\mathrm{tr}\bigl((I_n-H)^2\bigr)
=
\mathrm{tr}(I_n)-2\mathrm{tr}(H)+\mathrm{tr}(H^2)
=
n-2\mathrm{tr}(H)+\mathrm{tr}(H^2).
\]
Therefore
\begin{align*}
\mathbb{E}\bigl[\mathrm{Crit}_n(\lambda)\mid \Phi\bigr]
&=
\frac{1}{n}\|(I_n-H)m_n\|_2^2
+
\frac{\sigma^2}{n}\Bigl(n-2\mathrm{tr}(H)+\mathrm{tr}(H^2)\Bigr)
+
\frac{2\sigma^2}{n}\mathrm{tr}(H)
-
\sigma^2 \\
&=
\frac{1}{n}\|(I_n-H)m_n\|_2^2
+
\frac{\sigma^2}{n}\mathrm{tr}(H^2).
\end{align*}

On the other hand,
\[
m_n-HY = (I_n-H)m_n - H\varepsilon,
\]
so
\begin{align*}
\mathbb{E}\bigl[\|m_n-HY\|_2^2 \mid \Phi\bigr]
&=
\|(I_n-H)m_n\|_2^2
+
\mathbb{E}\bigl[\varepsilon^\top H^\top H \varepsilon \mid \Phi\bigr].
\end{align*}
Again using symmetry of $H$ and the conditional covariance of $\varepsilon$,
\[
\mathbb{E}\bigl[\varepsilon^\top H^\top H \varepsilon \mid \Phi\bigr]
=
\sigma^2 \mathrm{tr}(H^2).
\]
Therefore
\[
\mathbb{E}\bigl[\|m_n-HY\|_2^2 \mid \Phi\bigr]
=
\|(I_n-H)m_n\|_2^2 + \sigma^2 \mathrm{tr}(H^2).
\]
Dividing by $n$ yields
\[
\mathbb{E}\bigl[\mathrm{Crit}_n(\lambda)\mid \Phi\bigr]
=
\frac{1}{n}\mathbb{E}\bigl[\|m_n-HY\|_2^2 \mid \Phi\bigr]
=
P_n(\lambda).
\]
This proves the conditional unbiasedness of $\mathrm{Crit}_n(\lambda)$.

\subsection{Oracle inequality for criterion-based selection}
\label{app:proofs_oracle_inequality}

Define
\[
\hat\lambda_n \in \arg\min_{\lambda\in\Gamma_n}\mathrm{Crit}_n(\lambda),
\]
and let
\[
\Delta_n
:=
\sup_{\lambda\in\Gamma_n}
\bigl|
\mathrm{Crit}_n(\lambda)-P_n(\lambda)
\bigr|.
\]
Let $\lambda_n^\star \in \arg\min_{\lambda\in\Gamma_n} P_n(\lambda)$ be any oracle minimizer over the finite candidate set.

By definition of $\Delta_n$,
\[
P_n(\hat\lambda_n)
\le
\mathrm{Crit}_n(\hat\lambda_n)+\Delta_n.
\]
Since $\hat\lambda_n$ minimizes the empirical criterion over $\Gamma_n$,
\[
\mathrm{Crit}_n(\hat\lambda_n)
\le
\mathrm{Crit}_n(\lambda_n^\star).
\]
Applying the definition of $\Delta_n$ once more gives
\[
\mathrm{Crit}_n(\lambda_n^\star)
\le
P_n(\lambda_n^\star)+\Delta_n.
\]
Combining the three displays yields
\[
P_n(\hat\lambda_n)
\le
P_n(\lambda_n^\star)+2\Delta_n
=
\inf_{\lambda\in\Gamma_n}P_n(\lambda)+2\Delta_n.
\]
This proves the conditional oracle inequality
\[
P_n(\hat\lambda_n)
\le
\inf_{\lambda\in\Gamma_n}P_n(\lambda)+2\Delta_n.
\]

Taking conditional expectations with respect to the response noise while holding $\Phi$ fixed, we further obtain
\[
\mathbb{E}\bigl[P_n(\hat\lambda_n)\mid \Phi\bigr]
\le
\inf_{\lambda\in\Gamma_n}P_n(\lambda)
+
2\,\mathbb{E}\bigl[\Delta_n\mid \Phi\bigr].
\]
Thus the selected adjustment is as good as the best candidate in the discrete negative-capable family, up to twice the uniform criterion-estimation error.

\subsection{Negative-adjustment recovery and discrete-grid approximation}
\label{app:proofs_sign_recovery_grid}

We now show that, under a strict margin condition, the selected parameter must lie in the negative region.

Recall the decomposition
\[
\Gamma_n^- := \Gamma_n \cap [-\tau_0+\tau,\,0),
\qquad
\Gamma_n^+ := \Gamma_n \cap [0,\,L],
\]
and assume that both sets are nonempty. Suppose
\[
\inf_{\lambda\in\Gamma_n^+}P_n(\lambda)
-
\inf_{\lambda\in\Gamma_n^-}P_n(\lambda)
>
2\Delta_n.
\]
We claim that this implies
\[
\hat\lambda_n \in \Gamma_n^-.
\]

Assume for contradiction that $\hat\lambda_n \in \Gamma_n^+$. Then
\[
P_n(\hat\lambda_n)\ge \inf_{\lambda\in\Gamma_n^+}P_n(\lambda).
\]
On the other hand, by the oracle inequality proved above,
\[
P_n(\hat\lambda_n)
\le
\inf_{\lambda\in\Gamma_n}P_n(\lambda)+2\Delta_n
\le
\inf_{\lambda\in\Gamma_n^-}P_n(\lambda)+2\Delta_n.
\]
Hence
\[
\inf_{\lambda\in\Gamma_n^+}P_n(\lambda)
\le
\inf_{\lambda\in\Gamma_n^-}P_n(\lambda)+2\Delta_n,
\]
which contradicts the assumed strict margin condition. Therefore
\[
\hat\lambda_n\notin \Gamma_n^+.
\]
Since $\hat\lambda_n\in\Gamma_n^- \cup \Gamma_n^+$ by construction, it follows that
\[
\hat\lambda_n\in\Gamma_n^-.
\]
Thus a sufficiently strong negative-region advantage is necessarily recovered by the criterion-based selector.

We next connect the finite candidate family to the corresponding continuous search interval. Let
\[
\Lambda_n^{\mathrm{cont}} := [-\tau_0+\tau,\,L],
\]
and suppose the finite grid $\Gamma_n\subset \Lambda_n^{\mathrm{cont}}$ has mesh width
\[
h_n
:=
\sup_{\lambda\in\Lambda_n^{\mathrm{cont}}}
\min_{\tilde\lambda\in\Gamma_n}
|\lambda-\tilde\lambda|.
\]
Assume moreover that there exists a finite constant $C_n$ such that
\[
|P_n(\lambda_1)-P_n(\lambda_2)|
\le
C_n |\lambda_1-\lambda_2|
\qquad
\text{for all }\lambda_1,\lambda_2\in\Lambda_n^{\mathrm{cont}}.
\]
Let $\lambda_{n,\mathrm{cont}}^\star$ be any minimizer of $P_n(\lambda)$ over $\Lambda_n^{\mathrm{cont}}$. By definition of the mesh width, there exists $\tilde\lambda\in\Gamma_n$ such that
\[
|\tilde\lambda-\lambda_{n,\mathrm{cont}}^\star|\le h_n.
\]
By the Lipschitz assumption,
\[
P_n(\tilde\lambda)
\le
P_n(\lambda_{n,\mathrm{cont}}^\star)+C_n h_n
=
\inf_{\lambda\in\Lambda_n^{\mathrm{cont}}}P_n(\lambda)+C_n h_n.
\]
Since
\[
\inf_{\lambda\in\Gamma_n}P_n(\lambda)\le P_n(\tilde\lambda),
\]
we obtain
\[
\inf_{\lambda\in\Gamma_n}P_n(\lambda)
\le
\inf_{\lambda\in\Lambda_n^{\mathrm{cont}}}P_n(\lambda)+C_n h_n.
\]

Combining this approximation bound with the oracle inequality yields
\[
P_n(\hat\lambda_n)
\le
\inf_{\lambda\in\Lambda_n^{\mathrm{cont}}}P_n(\lambda)
+
C_n h_n
+
2\Delta_n.
\]
Therefore the data-driven selector over the discrete negative-capable family incurs only two explicit approximation losses relative to the continuous oracle: the grid discretization error $C_n h_n$ and the empirical criterion error $2\Delta_n$.

Taken together, these results justify the automatic-selection procedure used in the main text. The SURE-type criterion is conditionally unbiased, the selected parameter enjoys a finite-family oracle inequality, a sufficiently favorable negative-region margin forces negative selection, and a fine enough candidate grid tracks the continuous negative-capable oracle up to a controlled discretization term.

\end{document}